\documentclass[journal,twoside,web]{ieeecolor}
\usepackage{generic}
\usepackage{cite}
\usepackage{amsmath,amssymb,amsfonts}
\usepackage{graphicx}
\usepackage{hyperref}
\hypersetup{hidelinks=true}
\usepackage{textcomp}
\usepackage{graphicx}
\usepackage{epsfig}  
\usepackage{wrapfig}
\usepackage{indentfirst}  
\usepackage{color}
\usepackage[dvipsnames]{xcolor}
\usepackage{subcaption}
\usepackage{tikz}
\usepackage{soul}
\usepackage{xcolor}
\usepackage{mathtools}
\usepackage{optidef}
\usepackage{breqn}
\usepackage[linesnumbered,ruled,vlined]{algorithm2e}
\SetKwInput{kwAdd}{Add}
\SetKwInput{kwSet}{Set}
\SetKwInput{kwReset}{Reset}
\SetKwInput{kwUpdate}{Update}
\SetKwInput{kwClear}{Clear}
\SetKwInput{kwSave}{Save}
\SetKwInput{kwBuild}{Build}
\SetKwInput{kwIni}{Initialization}
\usetikzlibrary{matrix,chains,positioning,decorations.pathreplacing,arrows}

\newtheorem{definition}{Definition}
\newtheorem{lemma}{Lemma}
\newtheorem{remark}{Remark}
\newtheorem{theorem}{Theorem}
\newtheorem{assumption}{Assumption}

\newcommand{\bG}{{\mathbf G}}

\newcommand{\bX}{{\mathbf X}}

\newcommand{\bL}{{\mathbf L}}

\def\BibTeX{{\rm B\kern-.05em{\sc i\kern-.025em b}\kern-.08em
    T\kern-.1667em\lower.7ex\hbox{E}\kern-.125emX}}
\begin{document}
\title{Optimal Control of Nonlinear Systems with Unknown Dynamics}
\author{%
Wenjian Hao, Paulo C. Heredia, Shaoshuai Mou \thanks{W. Hao, P. Heredia, and S. Mou are with the School of Aeronautics
and Astronautics, Purdue University, West Lafayette, IN, 47906, USA. Email: \{hao93, pheredia, mous\}@purdue.edu.}}
\maketitle

\begin{abstract}
This paper presents a data-driven method for finding a closed-loop optimal controller, which minimizes a specified infinite-horizon cost function for systems with unknown dynamics given any arbitrary initial state. Suppose the closed-loop optimal controller can be parameterized by a given class of functions, hereafter referred to as the policy. The proposed method introduces a novel gradient estimation framework, which approximates the gradient of the cost function with respect to the policy parameters via integrating the Koopman operator with the classical concept of actor-critic. This enables the policy parameters to be tuned iteratively using gradient descent to achieve an optimal controller, leveraging the linearity of the Koopman operator. The convergence analysis of the proposed framework is provided. The effectiveness of the method is demonstrated through comparisons with a model-free reinforcement learning approach, and its control performance is further evaluated through simulations against model-based optimal control methods that solve the same optimal control problem utilizing the exact system dynamics.
\end{abstract}


\section{Introduction}
\label{sec:introduction}
\IEEEPARstart{O}{ptimal} control theory provides a foundational mathematical framework for the design of control strategies aimed at minimizing user-defined cost functions, typically under the assumption of known system dynamics \cite{anderson2007optimal}. For systems with unknown dynamics, model-free reinforcement learning (RL) \cite{sutton2018reinforcement} has emerged as a promising alternative to derive closed-loop optimal controllers directly from data. In RL, the unknown dynamics are modeled as a Markov decision process, and parameterized policies are employed as closed-loop controllers. A widely adopted structure is the actor–critic framework \cite{konda1999actor}, where the critic evaluates the policy using observed data and the actor updates the policy parameters based on the critic’s feedback. The critic approximates the expected cost by minimizing the temporal-difference error \cite{sutton1988learning} without requiring explicit knowledge of the system dynamics, while the actor improves the policy by optimizing against the critic. Prominent advances in RL include deep Q networks \cite{mnih2015human}, proximal policy optimization \cite{schulman2017proximal}, and
 deterministic policy gradient algorithms \cite{lillicrap2015continuous}, which extend Q-learning ideas to continuous action spaces \cite{silver2014deterministic}. Despite these successes, RL methods typically demand a large number of trial-and-error interactions to identify near-optimal controllers \cite{xiong2022deterministic}, posing significant challenges for real-world deployment.

To improve the efficiency of model-free methods in identifying closed-loop optimal controllers, recent attention has been given to data-driven model-based methods, which involve estimating system dynamics and subsequently deriving closed-loop optimal controllers using the learned models. For example, integral RL \cite{lewis2012optimal} directly determines optimal feedback gains from input-output data collected along system trajectories, typically assuming linear system dynamics. For systems with completely unknown nonlinear dynamics, several studies \cite{hernandaz1990neural,miller1990real,draeger1995model, chua2018deep, askari2025model} focus on learning unknown dynamics using neural networks, followed by leveraging classical model-based control strategies such as model predictive control (MPC) to compute optimal control actions. In the area of model-based RL \cite{doya2000reinforcement}, popular methods \cite{deisenroth2011pilco, chebotar2017combining} leverage Gaussian processes to approximate system dynamics, while works \cite{deisenroth2011learning,levine2016end} utilize deep neural networks (DNNs) to learn the dynamics. These techniques facilitate the generation of closed-loop optimal controllers by performing policy gradient updates using the learned models to directly minimize the user-defined cost function. Despite the successful integration of dynamic learning and policy design, several challenges persist, particularly due to the nonlinearity of learned dynamic models. These challenges include difficulties in verifying critical properties of the learned dynamics, such as controllability and stability, as noted in \cite{tanaka1996approach}. Furthermore, the significant nonlinearity in the learned dynamics often results in substantial computational complexity, which complicates practical implementation \cite{polydoros2017survey}.

The Koopman operator \cite{proctor2016dynamic} offers an alternative approach to approximate nonlinear systems with linear dynamics based on state-control pairs \cite{mezic2015applications, proctor2018generalizing, mauroy2016linear}, enabling the evaluation of key properties of the learned dynamics, such as observability and controllability. Techniques like extended dynamic mode decomposition \cite{korda2018convergence} transform the state space into a higher-dimensional space, making the dynamics approximately linear through carefully selected lifting functions. Operator-theoretic methods have further applied Koopman-based lifting of Hamiltonian systems via the Pontryagin maximum principle \cite{villanueva2021towards}. To address the complexity of choosing lifting functions, recent studies, for instance, \cite{lusch2017data} has employed deep learning techniques to discover the eigenfunctions of the Koopman operator. Deep Koopman operator (DKO) methods \cite{korda2018linear, dk, morton2018deep, hao2024deep} utilize DNNs as lifting functions, optimizing them with state-control pairs and a properly defined loss function. Model-based controllers, such as MPC, can then be integrated with the learned Koopman dynamics model \cite{korda2018linear}. Nevertheless, inaccuracies in the learned model may result in cumulative errors during cost function computation when propagating system states, thereby affecting the reliability of the approach. 

Motivated by the above, this paper proposes a data-driven framework to derive the closed-loop optimal controller via the combined benefits of actor-critic concepts and dynamics learning using the DKO approach. This is achieved by decomposing the bilevel optimization problem of the actor-critic structure through the integration of the DKO, enabling independent tuning of the dynamics approximator and the critic. The proposed approach is distinct from recent advancements in integrating the Koopman operator into RL. For instance, \cite{weissenbacher2022koopman} employs the Koopman operator to augment the static offline datasets during training, whereas the present work focuses on estimating the policy gradient using the Koopman operator. Similarly, \cite{rozwood2024koopman} leverages the Koopman operator to directly approximate the critic, enhancing the actor-critic framework, while the proposed method primarily leverages the DKO to predict future system states, facilitating improved policy optimization. The main contributions are summarized as follows: \begin{itemize}
\item To accelerate dynamics learning, the dynamics identification problem is formulated as a four-variable optimization problem. A multivariable update rule grounded in DKO is developed, and its convergence is analyzed. The results demonstrate faster convergence compared with standard single-parameter updates.
\item To mitigate model error accumulation during state propagation when computing the infinite-horizon cost gradient with respect to policy parameters, a DKO-based policy gradient estimation scheme is introduced. By embedding DKO learning into the actor–critic framework, the proposed method enables concurrent dynamics learning, critic refinement, and policy optimization using only one-step predictions from the DKO dynamics. 
\end{itemize}

The remainder of the paper is organized as follows. Section \ref{ProF} formulates the problem. Section \ref{proposedalg} introduces the proposed framework and offers the corresponding theoretical analysis. Section \ref{NSim} details an online algorithm for efficient framework implementation and validates the framework and theoretical findings through numerical simulations. Finally, Section \ref{Conc} provides concluding remarks.

\textbf{\emph{Notations.}} Let $\parallel \cdot \parallel$ be the Euclidean norm. For a matrix $A\in\mathbb{R}^{n\times m}$, $\parallel A \parallel_F$ denotes its Frobenius norm, $A'$ denotes its transpose, $A^\dagger$ denotes its Moore-Penrose pseudoinverse, $Tr(A)$ denotes its trace, and $\lambda_{\mathrm{min}}(AA')$ is the minimum eigenvalue of $AA'$. $\langle \cdot, \cdot \rangle$ denotes the inner product. Given an arbitrary function $\boldsymbol{f}(\boldsymbol{x}, \boldsymbol{y})$, $\nabla_{\boldsymbol{x}} \boldsymbol{f}(\boldsymbol{x}_k)\coloneqq \frac{\partial \boldsymbol{f}(\boldsymbol{x},\boldsymbol{y})}{\partial \boldsymbol{x}}\Bigr|_{\substack{\boldsymbol{x}_k}}$ and $\nabla_{\boldsymbol{x}\boldsymbol{x}} \boldsymbol{f}(\boldsymbol{x}_k)\coloneqq \frac{\partial^2 \boldsymbol{f}(\boldsymbol{x},\boldsymbol{y})}{\partial \boldsymbol{x}\partial \boldsymbol{x}}\Bigr|_{\substack{\boldsymbol{x}_k}}$ denote the first-order and second-order partial derivative of $\boldsymbol{f}(\boldsymbol{x}, \boldsymbol{y})$ with respect to $\boldsymbol{x}$ evaluated at $\boldsymbol{x}_k$ respectively, and $\nabla_{\boldsymbol{x}\boldsymbol{y}} \boldsymbol{f}(\boldsymbol{x}_k, \boldsymbol{y}_k)\coloneqq \frac{\partial^2 \boldsymbol{f}(\boldsymbol{x},\boldsymbol{y})}{\partial \boldsymbol{x} \partial \boldsymbol{y}}\Bigr|_{\substack{\boldsymbol{x}_k, \boldsymbol{y}_k}}$ denotes the second-order derivative of $\boldsymbol{f}(\boldsymbol{x},\boldsymbol{y})$ evaluated at $(\boldsymbol{x}_k, \boldsymbol{y}_k)$.

\section{The Problem}\label{ProF}
Consider the following discrete-time dynamical system:
\begin{equation}\label{eq_unknown_dyn} \boldsymbol{x}(t+1) = \boldsymbol{f}(\boldsymbol{x}(t),\boldsymbol{u}(t)),
\end{equation}
where $t = 0, 1, 2, \cdots$ denotes the time index, $\boldsymbol{x}(t)\in\mathcal{X}\subset\mathbb{R}^n$ and $\boldsymbol{u}(t)\in\mathcal{U} \subset\mathbb{R}^m$ denote the system state and control input at time $t$, respectively, and $\boldsymbol{f}: \mathcal{X}\times\mathcal{U}\rightarrow\mathcal{X}$ denotes the time-invariant, Lipschitz continuous, and unknown dynamics mapping. We assume $\mathcal{X}$ is a countably infinite state space. Notably, for the remainder of this paper, we distinguish between system state-input variables and fixed observed state-input pairs by denoting $\boldsymbol{x}_t$ and $\boldsymbol{u}_t$ as constant state and input vectors observed at time $t$, respectively. 

At any time $t^*\in\{0,1,2,\cdots \}$, let $\boldsymbol{x}(t^*) = \boldsymbol{x}_{t^*}$, an infinite-horizon cost function under the unknown dynamics in \eqref{eq_unknown_dyn} is defined as follows:
\begin{equation}\label{eq_inf_cosfunc}
    J_{t^*} = \sum_{t=t^*}^{\infty} \gamma^{t-t^*} c(\boldsymbol{x}(t), \boldsymbol{u}(t)),
\end{equation}
where $c: \mathcal{X}\times\mathcal{U}\rightarrow\mathbb{R}$ denotes a bounded and Lipschitz continuous stage cost function, $0<\gamma< 1$ is a discount factor, and the control inputs are assumed to be chosen from a policy obeying the following form:
\begin{equation}\label{eq_policy}
    \boldsymbol{u}(t) = \boldsymbol{\mu}(\boldsymbol{x}(t), \boldsymbol{\theta}^\mu),
\end{equation} where $\boldsymbol{\mu}(\cdot, \boldsymbol{\theta}^\mu):\mathcal{X}\rightarrow\mathcal{U}$ denotes a known Lipschitz continuous function with a tunable parameter $\boldsymbol{\theta}^\mu\in\mathbb{R}^q$.  

Assume that the closed-loop optimal controller that minimizes $J_{t^*}$ in \eqref{eq_inf_cosfunc} can be represented by $\boldsymbol{\mu}(\boldsymbol{x}(t), \boldsymbol{\theta}^{\mu *})$. The \textbf{problem of interest} is to develop a data-driven framework to achieve $\boldsymbol{\theta}^{\mu *}$. Specifically, given an arbitrary initial state $\boldsymbol{x}(t^*) = \boldsymbol{x}_{t^*}$ at time $t^*$, this paper aims to solve the following optimization problem without knowing the actual dynamics $\boldsymbol{f}$ in \eqref{eq_unknown_dyn}:
\begin{equation}\label{eq_pf_oc}
\begin{aligned}
      \boldsymbol{\theta}^{\mu *} &= \arg\min_{\boldsymbol{\theta}^\mu\in\mathbb{R}^q} J_{t^*}(\boldsymbol{\theta}^\mu) \\ \quad
 \text{subject to:} \quad \boldsymbol{x}(t+1) &= \boldsymbol{f}(\boldsymbol{x}(t),\boldsymbol{\mu}(\boldsymbol{x}(t), \boldsymbol{\theta}^\mu)), \boldsymbol{f}\ \text{unknown},\\  \boldsymbol{x}(t^*) &= \boldsymbol{x}_{t^*}, \quad \boldsymbol{x}_{t^*}\ \text{given}.
\end{aligned}
\end{equation}
\begin{remark}
The optimization problem in \eqref{eq_pf_oc} is a well-established optimal control problem when the dynamics $\boldsymbol{f}$ in \eqref{eq_unknown_dyn} is known \cite{lehtomaki1981robustness}. In the case where $\boldsymbol{f}$ is unknown, one approach to solve \eqref{eq_pf_oc} involves model-free methods, such as RL techniques, which typically require a substantial amount of data to achieve an optimal solution. In the following section, we will propose a framework that concurrently approximates $\boldsymbol{f}$ using the Koopman operator and optimizes $\boldsymbol{\theta}^\mu$ to find the $\boldsymbol{\theta}^{\mu *}$. The method is shown to have better efficiency of computing $\boldsymbol{\theta}^{\mu *}$ in later simulations.    
\end{remark}

\section{Main Results}\label{proposedalg}
In this section, we first identify the primary challenges and fundamental concepts underlying the proposed framework. Afterward, we introduce a data-driven tuning framework to solve \eqref{eq_pf_oc}. Finally, we present the convergence analysis of the proposed framework.
\subsection{Challenges and Key Ideas}
An iterative method to find $\boldsymbol{\theta}^{\mu*}$ in \eqref{eq_pf_oc} is to employ the gradient descent method. Specifically, let $k=0,1,2,\cdots$ denote the iteration index, $\boldsymbol{\theta}_k^\mu\in\mathbb{R}^q$ denote the estimation of $\boldsymbol{\theta}^{\mu*}$ at the $k$-th iteration, and $\alpha_k^\mu$ represent the step size at the $k$-th iteration corresponding to $ \boldsymbol{\theta}_k^\mu$. Given any arbitrary initial $\boldsymbol{\theta}_0^\mu$, the parameter $\boldsymbol{\theta}_k^\mu$ is updated iteratively using the gradient of $J_{t^*}$ as follows:
\begin{equation}\label{eq_gd_mu}
    \boldsymbol{\theta}_{k+1}^\mu = \boldsymbol{\theta}_k^\mu - \alpha_k^\mu \nabla_{\boldsymbol{\theta}^\mu}J_{t^*} (\boldsymbol{\theta}_k^\mu), \quad \boldsymbol{\theta}_0^\mu\ \text{given},
\end{equation}
where 
\begin{equation}\label{eq_pg}
\begin{aligned}
\nabla_{\boldsymbol{\theta}^\mu}J_{t^*}
= \frac{\partial c_{t^*}}{\partial \boldsymbol{u}_{t^*}}\frac{\partial \boldsymbol{u}_{t^*}}{\partial \boldsymbol{\theta}^\mu}+ \sum_{t=t^*+1}^\infty \gamma^{t-t^*}(\frac{\partial c_t}{\partial \boldsymbol{u}(t)} \frac{\partial \boldsymbol{u}(t)}{\partial \boldsymbol{\theta}^\mu} \\+ \frac{\partial c_t}{\partial \boldsymbol{x}(t)} \frac{\partial \boldsymbol{x}(t)}{\partial \boldsymbol{u}(t-1)} \frac{\partial \boldsymbol{u}(t-1)}{\partial \boldsymbol{\theta}^\mu}).
\end{aligned}
\end{equation}
Here, $c_t \coloneqq c(\boldsymbol{x}(t), \boldsymbol{u}(t))$ is introduced for notation brevity, and the selection of $\alpha_k^\mu$ will be discussed in detail later in this paper. Throughout this paper, we refer to $\nabla_{\boldsymbol{\theta}^\mu}J_{t^*}$ as the policy gradient.

Two key challenges arise when computing the policy gradient $\nabla_{\boldsymbol{\theta}^\mu}J_{t^*}$ in \eqref{eq_pg} due to the unknown dynamics $f$ in \eqref{eq_unknown_dyn}. First, the system dynamics $\boldsymbol{f}$ is assumed to be unknown in the present work, making the gradient $\frac{\partial \boldsymbol{x}(t)}{\partial \boldsymbol{u}(t-1)}$ unknown as well. Second, calculating $\sum_{t=t^*+1}^\infty\frac{\partial c_t}{\partial \boldsymbol{x}(t)}\frac{\partial \boldsymbol{x}(t)}{\partial \boldsymbol{u}(t-1)}\frac{\partial \boldsymbol{u}(t-1)}{\partial \boldsymbol{\theta}^\mu}+\frac{\partial c_t}{\partial \boldsymbol{u}(t)} \frac{\partial \boldsymbol{u}(t)}{\partial \boldsymbol{\theta}^\mu}$ requires future system states $\boldsymbol{x}(t^*+1), \boldsymbol{x}(t^*+2), \cdots$ that evolve according to system dynamics. Even if one can approximate the dynamics $\boldsymbol{f}$ with small estimation errors, these errors may still accumulate over an infinite time horizon, potentially resulting in inaccuracies in calculating $\nabla_{\boldsymbol{\theta}^\mu} J_{t^*}$. 

To address the two challenges in computing the policy gradient, we propose a framework that applies the deep Koopman operator (DKO) to approximate the unknown dynamics $\boldsymbol{f}$ in a linear form, enabling efficient approximation of $\frac{\partial \boldsymbol{x}(t+1)}{\partial \boldsymbol{u}(t)}$ and potentially improving gradient-based optimization. Subsequently, we use the temporal difference error technique \cite{sutton1988learning} to approximate $J_{t^*+1}$ such that $\nabla_{\boldsymbol{\theta}^\mu} J_{t^*+1}$ can be estimated using only a one-time-step prediction from the DKO to reduce the computational complexity.

\subsection{The Proposed Framework}
We now introduce a data-driven framework to approximate the policy gradient $\nabla_{\boldsymbol{\theta}^\mu}J_{t^*}$ in \eqref{eq_pg} such that one can solve \eqref{eq_pf_oc} by tuning $\boldsymbol{\theta}^\mu$ following \eqref{eq_gd_mu}. The proposed gradient estimation framework comprises three components, updated concurrently: (i) a DKO block for approximating the unknown system dynamics and enabling future state predictions; (ii) a critic block for estimating the cost function $J_{t^*}$ in \eqref{eq_inf_cosfunc}; and (iii) an actor block for optimizing the policy based on the critic's evaluation, utilizing a one-time-step prediction from the DKO.

To proceed, we consider a given dataset consisting of all observed tuples, represented as $\mathcal{D} = \cup_{i=1}^{N} \{(\boldsymbol{x}_i, \boldsymbol{u}_i, \boldsymbol{x}_i^+)\}$, with its index set denoted by $\mathcal{I}_D = \{1,2,\cdots,N\}$. The dataset $\mathcal{D}$ is not restricted to a specific collection procedure. In practice, the input $\boldsymbol{u}_i$ can be sampled from some probability distribution with non-zero probability over all actions (i.e., a behavior policy) or generated by a human operator. Here, $\boldsymbol{x}_i \coloneqq \boldsymbol{x}_{t_i}$ and $\boldsymbol{u}_i \coloneqq \boldsymbol{u}_{t_i}$ represent the system state and control input observed at time $t_i$, respectively, while $\boldsymbol{x}_i^+ \coloneqq \boldsymbol{x}_{t_i+1}$ denotes the resulting system state obtained from $\boldsymbol{x}_i$ after applying $\boldsymbol{u}_i$ to the unknown dynamics in \eqref{eq_unknown_dyn}. This notation is introduced because the proposed framework does not require the tuples to be sequential, i.e., $t_i +1$ is not necessarily equal to $t_{i+1}$. 

\textbf{Dynamics approximation using DKO.} To approximate the unknown dynamics $\boldsymbol{f}$ in \eqref{eq_unknown_dyn}, this paper aims to obtain the following estimated dynamics:  \begin{equation}\label{eq_approx_f}
\begin{aligned}
    \boldsymbol{x}(t+1) = C^*\Big(A^* \boldsymbol{g}(\boldsymbol{x}(t),\boldsymbol{\theta}^{f*}) + B^* \boldsymbol{u}(t)\Big),
\end{aligned}
\end{equation} where $\boldsymbol{g}(\boldsymbol{x},\boldsymbol{\theta}^{f*}) = \boldsymbol{\phi}_f(\boldsymbol{x})\boldsymbol{\theta}^{f*}$, and the feature function $\boldsymbol{\phi}_f(\boldsymbol{x}): \mathcal{X}\rightarrow\mathbb{R}^{r\times p}$ satisfies  $\parallel\boldsymbol{\phi}_f(\boldsymbol{x}) \parallel\leq 1$ and $r\geq n$.  $\boldsymbol{\theta}^{f*}\in\mathbb{R}^p$ and $A^*\in\mathbb{R}^{r\times r}, B^*\in\mathbb{R}^{r\times m}, C^*\in\mathbb{R}^{n\times r}$ are the constant parameter vector and matrices to be determined, respectively. The function $\boldsymbol{g}$ is assumed to be Lipschitz continuous. Here, \eqref{eq_approx_f} is derived based on the following DKO representation:
\begin{align}
    \boldsymbol{g}(\boldsymbol{x}(t+1),\boldsymbol{\theta}^{f*}) &= A^* \boldsymbol{g}(\boldsymbol{x}(t),\boldsymbol{\theta}^{f*}) + B^* \boldsymbol{u}(t), \label{eq_lift_koopman} \\
\boldsymbol{x}(t+1) &= C^* \boldsymbol{g}(\boldsymbol{x}(t+1),\boldsymbol{\theta}^{f*}), \label{eq_x_koopman}
\end{align}
where \eqref{eq_lift_koopman} describes the evolution of the system dynamics in the lifted space and \eqref{eq_x_koopman} additionally assumes the existence of a linear mapping between $\boldsymbol{x}(t+1)$ and its lifted states $\boldsymbol{g}(\boldsymbol{x}(t+1),\boldsymbol{\theta}^{f *})$. We refer to \cite{hao2024deep} for a detailed analysis of the estimation errors of \eqref{eq_approx_f} with respect to the structure of $\boldsymbol{g}$. 

To achieve \eqref{eq_approx_f}, we formulate the following optimization problem using $\mathcal{D}$: \[A^*, B^*, C^*, \boldsymbol{\theta}^{f*} = \arg\min_{A,B,C,\boldsymbol{\theta}^f}\bL_f(A,B,C,\boldsymbol{\theta}^f),\] 
where
\begin{equation}\label{eq_L_f}
\begin{aligned}
        \bL_f =  \frac{1}{2N}\sum_{i\in\mathcal{I}_D}(\underbrace{\parallel\boldsymbol{g}(\boldsymbol{x}_i^+,\boldsymbol{\theta}^f) -A\boldsymbol{g}(\boldsymbol{x}_i,\boldsymbol{\theta}^f) -B\boldsymbol{u}_i \parallel^2}_{\delta_i(A,B,\boldsymbol{\theta}^f)} \\+ \underbrace{\parallel \boldsymbol{x}_i^+ - C\boldsymbol{g}(\boldsymbol{x}_i^+,\boldsymbol{\theta}^f) \parallel^2}_{\bar{\delta}_i(C,\boldsymbol{\theta}^f)}) 
    \end{aligned}
\end{equation} with $\mathcal{I}_D = \{1,2,\cdots,N\}$ the index set of $\mathcal{D}$. Here, $\delta_i(A,B,\boldsymbol{\theta}^f)$ and $\bar{\delta}_i(C,\boldsymbol{\theta}^f)$ are designed to approximate \eqref{eq_lift_koopman} and \eqref{eq_x_koopman}, respectively. Let $\boldsymbol{\theta}_k^f$ be the estimation of $\boldsymbol{\theta}^{f*}$ at iteration $k$. We construct the following data matrices from $\mathcal{D}$: \begin{equation}\label{xyudata}
    \begin{aligned}
    \bX &=[\boldsymbol{x}_1, \boldsymbol{x}_2,\cdots,\boldsymbol{x}_N] \in \mathbb{R}^{n \times N},\\
    \mathbf{U} &=[\boldsymbol{u}_1, \boldsymbol{u}_2,\cdots,\boldsymbol{u}_N]\in \mathbb{R}^{m \times N},\\
    \bar{\bX} &= [\boldsymbol{x}_1^+, \boldsymbol{x}_2^+,\cdots,\boldsymbol{x}_N^+]\in \mathbb{R}^{n \times N},\\ 
    \bG_k&= [\boldsymbol{g}(\boldsymbol{x}_1,\boldsymbol{\theta}_k^f),\boldsymbol{g}(\boldsymbol{x}_2,\boldsymbol{\theta}_k^f),\cdots, \boldsymbol{g}(\boldsymbol{x}_N, \boldsymbol{\theta}_k^f)]\in \mathbb{R}^{r \times N},\\
    \bar{\bG}_k&= [\boldsymbol{g}(\boldsymbol{x}_1^+,\boldsymbol{\theta}_k^f),\boldsymbol{g}(\boldsymbol{x}_2^+,\boldsymbol{\theta}_k^f),\cdots, \boldsymbol{g}(\boldsymbol{x}_N^+,\boldsymbol{\theta}_k^f)] \in \mathbb{R}^{r \times N}.
\end{aligned}
\end{equation}
If the matrices $\bar{\bG}_k \in \mathbb{R}^{r\times N}$ and $\begin{bmatrix} \bG_k \\ \mathbf{U} \end{bmatrix}\in \mathbb{R}^{(r+m)\times N}$ are with full row rank, meaning they are right-invertible, we propose the following updating rule to obtain \eqref{eq_approx_f}: 
\begin{equation}\label{eq_gd_thetaf}
    \boldsymbol{\theta}_{k+1}^f = \boldsymbol{\theta}_k^f -\alpha_k^f\nabla_{\boldsymbol{\theta}^f}\bL_f(A_k, B_k, C_k, \boldsymbol{\theta}_k^f), \quad \boldsymbol{\theta}_0^f\ \text{given},
\end{equation}
where 
\begin{equation}\label{lmn}
    \begin{aligned}
        [A_k\ B_k] &= \arg\min_{[A\ B]} \sum_{i\in\mathcal{I}_D}\delta_i(A,B,\boldsymbol{\theta}_k^f) = \bar{\bG}_k \begin{bmatrix} \bG_k \\ \mathbf{U} \end{bmatrix}^\dagger, \\
        C_k &= \arg\min_{C} \sum_{i\in\mathcal{I}_D}\bar{\delta}_i(C,\boldsymbol{\theta}_k^f)=  \bar{\bX}\bar{\bG}_k^{\dagger},
    \end{aligned}
\end{equation} are constant matrices determined by $\boldsymbol{\theta}_k^f$. In Section~\ref{analysis}, we present the convergence analysis of the update rule in \eqref{eq_gd_thetaf}.

\textbf{Cost function approximation (critic).} To approximate $J_{t^*}$ in \eqref{eq_inf_cosfunc} under a fixed policy, this paper employs a parameterized function of the form $J_{t^*}=V(\boldsymbol{x}_{t^*},\boldsymbol{\theta}^{J*})$, where $V(\boldsymbol{x},\boldsymbol{\theta}^{J*}) = \boldsymbol{\phi}_J(\boldsymbol{x})'\boldsymbol{\theta}^{J*}$ with the feature function $\boldsymbol{\phi}_J(\boldsymbol{x}): \mathcal{X}\rightarrow\mathbb{R}^s$ satisfying $\parallel\boldsymbol{\phi}_J(\boldsymbol{x}) \parallel\leq 1$, and $\boldsymbol{\theta}^{J*}\in\mathbb{R}^s$ denotes the optimal parameter vector to be identified. The function $V$ is assumed to be Lipschitz continuous and the matrix $[\boldsymbol{\phi}_J(\boldsymbol{x_1}), \boldsymbol{\phi}_J(\boldsymbol{x_2}),\cdots,\boldsymbol{\phi}_J(\boldsymbol{x}_N)]'\in\mathbb{R}^{N\times s}$ is assumed to have full column rank. $\boldsymbol{\theta}^{J*}$ is obtained by minimizing the temporal difference (TD) loss via gradient descent using $\mathcal{D}$ \cite{sutton1988learning}:
\begin{equation} \label{TD_update}
\boldsymbol{\theta}_{k+1}^J = \boldsymbol{\theta}_k^J -\alpha_k^J \nabla_{\boldsymbol{\theta}^J}\bL_J(\boldsymbol{\theta}_k^J), \quad \boldsymbol{\theta}_0^J\ \text{given},
\end{equation}
where \begin{equation}\label{eq_td_err}
\begin{aligned}
\bL_J(\boldsymbol{\theta}^J)= \frac{1}{2N}\sum_{i\in\mathcal{I}_D}(c(\boldsymbol{x}_i, \boldsymbol{u}_i) + \gamma V(\boldsymbol{x}_i^+,\boldsymbol{\theta}^J) \!-\! V(\boldsymbol{x}_i,\boldsymbol{\theta}^J))^2. \nonumber
\end{aligned}
\end{equation} Here, $\boldsymbol{\theta}_k^J$ is the estimation of $\boldsymbol{\theta}^{J*}$ at iteration $k$, and $\alpha_k^J$ is the corresponding step size.

\textbf{Policy update (actor).} To obtain $\boldsymbol{\theta}^{\mu *}$ that accounts for diverse initial states, after updating $\boldsymbol{\theta}_k^f$ and $\boldsymbol{\theta}_k^J$ following \eqref{eq_gd_thetaf} and \eqref{TD_update}, respectively, $\boldsymbol{\theta}_k^\mu$ is updated using the approximated policy gradient computed over $\mathcal{D}$ as follows:
\begin{equation}\label{eq_gd_L2}
\begin{aligned}
    \boldsymbol{\theta}_{k+1}^\mu = \boldsymbol{\theta}_k^\mu - \alpha_k^\mu \nabla_{\boldsymbol{\theta}^\mu}\hat{J} (\boldsymbol{\theta}_{k+1}^f, \boldsymbol{\theta}_{k+1}^J,\boldsymbol{\theta}_k^\mu), \quad \boldsymbol{\theta}_0^\mu\ \text{given},
\end{aligned}
\end{equation}
where
\begin{equation}\label{eq_approximate_pg}
\begin{aligned}
\nabla_{\boldsymbol{\theta}^\mu}\hat{J} = \frac{1}{N} \sum_{i\in\mathcal{I}_D} \Big(\nabla_{\boldsymbol{\mu}}c(\boldsymbol{x}_i, \boldsymbol{\mu}(\boldsymbol{x}_i, \boldsymbol{\theta}_k^\mu)) \nabla_{\boldsymbol{\theta}^\mu}\boldsymbol{\mu}(\boldsymbol{x}_i, \boldsymbol{\theta}_k^\mu) \\ + 
\gamma \nabla_{\boldsymbol{\hat x}}V(\boldsymbol{\hat{x}}_i^+, \boldsymbol{\theta}_{k+1}^J) \nabla_{\boldsymbol{\mu}}\boldsymbol{\hat{x}}_i^+ \nabla_{\boldsymbol{\theta}^\mu}\boldsymbol{\mu}(\boldsymbol{x}_i, \boldsymbol{\theta}_k^\mu)\Big)
\end{aligned}
\end{equation} with $\boldsymbol{\hat{x}}_i^+ = C_k(A_k \boldsymbol{g}(\boldsymbol{x}_i,\boldsymbol{\theta}_{k+1}^f) + B_k \boldsymbol{\mu}(\boldsymbol{x}_i, \boldsymbol{\theta}_k^\mu))$ the one-step predicted state from DKO and $\nabla_{\boldsymbol{\mu}}\boldsymbol{\hat{x}}_i^+ = C_kB_k$. Here, \eqref{eq_approximate_pg} approximates the policy gradient $\nabla_{\boldsymbol{\theta}^\mu}J_{t^*}$ in \eqref{eq_pg} at iteration $k$, where $\hat{J}=\frac{1}{N}\sum_{t^*\in\mathcal{I}_D}\hat{J}_{t^*}(\boldsymbol{\theta}_{k+1}^f, \boldsymbol{\theta}_{k+1}^J,\boldsymbol{\theta}_k^\mu)$. The term $\hat{J}_{t^*}$ is obtained by replacing the true dynamics and $J_{t^*+1}$ in \eqref{eq_inf_cosfunc}, i.e., $J_{t^*} = c(\boldsymbol{x}_{t^*},\boldsymbol{\mu}(\boldsymbol{x}_{t^*}, \boldsymbol{\theta}_k^\mu))+ 
\gamma J_{t^*+1}$, with the DKO dynamics and critic output, respectively, yielding
\begin{equation}\label{eq_J_hat}
    \hat{J}_{t^*} = c(\boldsymbol{x}_{t^*},\boldsymbol{\mu}(\boldsymbol{x}_{t^*}, \boldsymbol{\theta}_k^\mu)) + \gamma V(\boldsymbol{\hat{x}}_{t^*}^+, \boldsymbol{\theta}_{k+1}^J).
\end{equation}

\begin{figure}[ht]
    \centering
\includegraphics[width=0.85\linewidth]{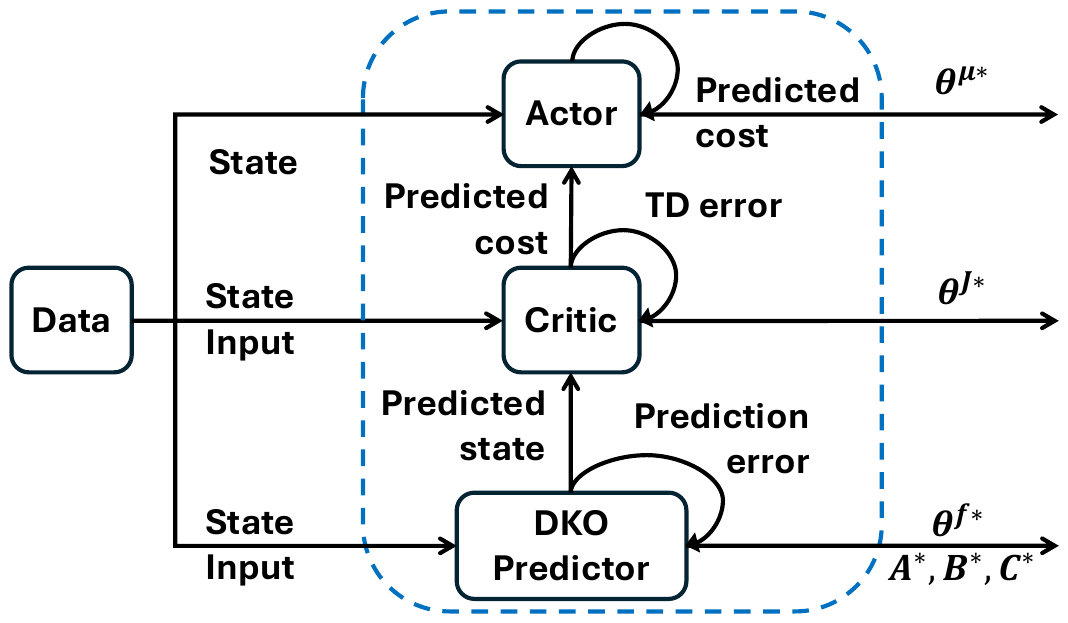}
    \caption{PGDK framework for policy gradient estimation.}
    \label{fig:alg_arch}
\end{figure}

To summarize, the proposed policy gradient approximation framework in \eqref{eq_approximate_pg} is referred to as the \textit{policy gradient with deep Koopman representation} (PGDK) throughout this paper, where the parameters $\boldsymbol{\theta}^f$ and $\boldsymbol{\theta}^J$ are updated according to \eqref{eq_gd_thetaf} and \eqref{TD_update}, respectively. The overall structure of PGDK is illustrated in Fig. \ref{fig:alg_arch}, while an offline implementation of this framework is provided in Algorithm~\ref{alg}.
\begin{algorithm}[h]
\SetAlgoLined
\LinesNumbered
\textbf{Input:} Dataset $\mathcal{D}$. \\ \textbf{Initialization:} Initialize $\boldsymbol{\theta}_0^f\in\mathbb{R}^p$, $\boldsymbol{\theta}_0^J\in\mathbb{R}^s$, and $\boldsymbol{\theta}_0^\mu\in\mathbb{R}^q$ for $\boldsymbol{g}(\cdot,\boldsymbol{\theta}^f)$, $ V(\cdot,\boldsymbol{\theta}^J)$, and $\boldsymbol{\mu}(\cdot, \boldsymbol{\theta}^\mu)$, respectively.
Set step size sequences $\{\alpha_k^f\}_{k=0}^K$, $\{\alpha_k^J\}_{k=0}^K$, $\{\alpha_k^\mu\}_{k=0}^K$, and discount factor $\gamma$.\\
    \For{$k=0,1,\cdots,K$}{
       Compute $A_k$, $B_k$ and $C_k$ following \eqref{lmn}.\\
       Update $\boldsymbol{\theta}_k^f$ and $\boldsymbol{\theta}_k^J$ following \eqref{eq_gd_thetaf} and \eqref{TD_update}, respectively.\\
       Update $\boldsymbol{\theta}_k^\mu$ following \eqref{eq_gd_L2}.
        }
\caption{Policy Gradient with Deep Koopman Representation (PGDK) — offline implementation}\label{alg}
\end{algorithm}

\subsection{Analysis} \label{analysis}
In this subsection, we analyze the convergence of the proposed framework in \eqref{eq_gd_L2} for a given dataset $\mathcal{D}$. To this end, we first introduce the following assumption and definition:
\begin{assumption}\label{asp1}
    The unknown dynamics $\boldsymbol{f}$ in \eqref{eq_unknown_dyn} can be rewritten as a deep Koopman dynamics in \eqref{eq_approx_f}, and $J_{t^*}$ in \eqref{eq_inf_cosfunc} can be expressed by the parameterized function $V(\boldsymbol{x}_{t^*},\boldsymbol{\theta}^{J*})$.
\end{assumption}
As noted in \cite{brunton2016koopman}, a finite-dimensional Koopman representation in \eqref{eq_lift_koopman} is achievable only if $\boldsymbol{f}$ has a single isolated fixed point.
\begin{definition}
    A sequence of step sizes $\{\alpha_k\}_{k=0}^\infty$ with $\alpha_k\geq 0$ is said to satisfy the Robbins–Monro (RM) condition if $\sum_{k=0}^\infty \alpha_k = \infty$ and $\sum_{k=0}^\infty \alpha_k^2 <\infty$.
\end{definition}

Since $\boldsymbol{\theta}^\mu$ is updated based on $\boldsymbol{\theta}^f$ and $\boldsymbol{\theta}^J$, it is first necessary to analyze the convergence of $\boldsymbol{\theta}_k^f$ in \eqref{eq_gd_thetaf} and $\boldsymbol{\theta}_k^J$ in \eqref{TD_update}. Thus, we establish the following key result regarding the dynamics approximation with respect to $\boldsymbol{\theta}^f$:
\begin{lemma}\label{thm1}
If Assumption~\ref{asp1} holds and $\boldsymbol{\theta}_k^f$ is updated following \eqref{eq_gd_thetaf} with step size $\alpha_k^f = \frac{1}{L_{f1}(2+k)}$, then \begin{equation}\label{eq_thm1_results}
    \parallel\boldsymbol{\theta}_k^f-\boldsymbol{\theta}^{f*}\parallel^2\leq \frac{\nu_f}{2+k},
\end{equation}
where $\nu_f = \max\{L_{f2}/L_{f1}^2, 2\parallel\boldsymbol{\theta}_0^f-\boldsymbol{\theta}^{f*}\parallel^2\}$, $L_{f1}$ is a constant, and $L_{f2}$ is a constant determined by the residual vectors from the least squares solutions in \eqref{lmn}. Furthermore, to guarantee $\lim_{k\rightarrow\infty}\parallel\boldsymbol{\theta}_k^f-\boldsymbol{\theta}^{f*}\parallel^2=0$, the step size sequence $\{\alpha_k^f\}_{k=0}^\infty$ must satisfy the RM condition.
\end{lemma} 
The proof of Lemma~\ref{thm1} is provided in the Appendix. Lemma~\ref{thm1} says that, under certain assumptions, the update rule in \eqref{eq_gd_thetaf} converges sublinearly. Following established TD-error results, we summarize from \cite{bhandari2018finite}:
\begin{lemma}\label{TD_result}[Theorem $1$, \cite{bhandari2018finite}]
    If the dynamics \eqref{eq_unknown_dyn} following the policy $\boldsymbol{\mu}$ is ergodic with a unique stationary distribution $\pi(\boldsymbol{x}):\mathbb{R}^n\rightarrow\mathbb{R}$. That is, for any two states $\boldsymbol{x}_1, \boldsymbol{x}_2: \pi(\boldsymbol{x}_2) = \lim_{t\rightarrow\infty}\mathbb{P}(\boldsymbol{x}_t = \boldsymbol{x}_2| \boldsymbol{x}_0 = \boldsymbol{x}_1)$. Let $\Sigma=\sum_{\boldsymbol{x}\in\mathcal{X}}\pi(\boldsymbol{x})\boldsymbol{\phi}_J(\boldsymbol{x})\boldsymbol{\phi}_J(\boldsymbol{x})'$, and denote $\omega_J$ as the smallest eigenvalue of $\Sigma$, and $\nu_J = (1-\gamma)^2\omega_J/4$. Given $\omega_J>0$ and a constant step size $\alpha_J= (1-\gamma)/4$, following \eqref{TD_update} yields: \[\parallel \boldsymbol{\theta}_k^J-\boldsymbol{\theta}^{J*}\parallel^2 \leq (1-\nu_J)^k\parallel\boldsymbol{\theta}_0^J-\boldsymbol{\theta}^{J*}\parallel^2.\]
\end{lemma}
Building on the preceding results, the approximated policy gradient satisfies the following convergence properties:
\begin{theorem}\label{thm2}
If Assumption~\ref{asp1} holds and the parameters $\boldsymbol{\theta}_k^f$, $\boldsymbol{\theta}_k^J$, and $\boldsymbol{\theta}_k^\mu$ are updated according to \eqref{eq_gd_thetaf}, \eqref{TD_update}, and \eqref{eq_gd_L2} with step sizes $\alpha_k^f = \frac{1}{L_{f1}(2+k)}$, $\alpha_J= (1-\gamma)/4$, and $\alpha_k^\mu = (k+1)^{-1/4}$, respectively. Then, the policy gradient satisfies
\begin{equation}
\begin{aligned}
&\min_{k\in\{0,1,2,\cdots,K-1\}}\parallel \nabla_{\boldsymbol{\theta}^\mu}\hat{J} (\boldsymbol{\theta}_{k+1}^f, \boldsymbol{\theta}_{k+1}^J,\boldsymbol{\theta}_k^\mu) \parallel^2 \\ \leq &L_a K^{-1/4} + L_b K^{-3/4} +L_c K^{-5/4}, \nonumber
\end{aligned}\end{equation}
where $L_a$, $\lim_{k\rightarrow\infty}L_b$, and $L_c$ are constants defined in the Appendix. The constant $L_{f1}$ is as specified in Lemma~\ref{thm1}. Moreover, to ensure $\lim_{k\rightarrow\infty}\parallel \nabla_{\boldsymbol{\theta}^\mu}\hat{J} (\boldsymbol{\theta}_{k+1}^f, \boldsymbol{\theta}_{k+1}^J, \boldsymbol{\theta}_k^\mu) \parallel^2 = 0$, $\{\alpha_k^\mu\}_{k=0}^\infty$ needs to satisfy the RM condition.
\end{theorem} The proof of Theorem~\ref{thm2} is provided in the Appendix. Theorem~\ref{thm2} establishes that the policy gradient converges when the policy is updated on a faster time scale than the DKO learner. Furthermore, it shows that the proposed method achieves a globally optimal policy provided that the functions $c$, $\boldsymbol{g}$, and $V$ are convex. In the non-convex case, the method converges only to a locally optimal policy.

If Assumption~\ref{asp1} does not hold, then $\nabla_{\boldsymbol{\theta}^\mu}\hat{J}$ in \eqref{eq_approximate_pg} only approximates $\nabla_{\boldsymbol{\theta}^\mu}J_{t^*}$ in \eqref{eq_pg} with a deterministic error $\epsilon_k$. Since $\boldsymbol{f}$, $J_{t^*}$, $\boldsymbol{g}$, and $V$ are Lipschitz continuous, there exists $\epsilon > 0$ such that $\parallel\epsilon_k\parallel\leq \epsilon$. According to \cite{bertsekas2000gradient}, if $J_{t^*}$ is strongly convex and a time-decaying step size $\alpha_k^\mu\rightarrow 0$ is employed, then $\lim_{k\rightarrow\infty}\parallel\boldsymbol{\theta}_k^\mu-\boldsymbol{\theta}^{\mu *}\parallel^2$ converges to a positive constant that is proportional to $\epsilon^2$.

\section{Numerical Simulations}\label{NSim}
In this section, we first present an online algorithm to implement the proposed framework efficiently. Then, we evaluate the performance of both the offline and online algorithms through numerical simulations.

\subsection{Online Implementation}
In practical scenarios, the offline approach presents challenges, particularly when $N$ is large, making dataset acquisition and gradient computation computationally intensive. Moreover, discrepancies between the current policy and the behavior policy employed during offline data collection may impede the identification of the closed-loop optimal controller, as noted in \cite{agarwal2020optimistic}. To address these limitations arising from training data constraints and to enhance implementation efficiency, we introduce an online variant of PGDK that extends Algorithm~\ref{alg}. In this formulation, instead of utilizing a fixed offline dataset $\mathcal{D}$ to estimate the policy gradient, the proposed online method employs the mini-batch gradient descent scheme \cite{mnih2013playing}, widely employed in RL. Specifically, at each iteration $k$, the policy parameters are updated using gradients computed from a sampled mini-batch, as described below.

\textbf{Data batches sampling.} To ensure the sufficient exploration of the system state space $\mathcal{X}$, we adopt the widely-used method from \cite{lillicrap2015continuous}, where an exploration policy is implemented by introducing noise $W(t)\in\mathcal{W}\subset\mathbb{R}^m$, sampled from a noise process, into the policy in \eqref{eq_policy} at each time step $t$ while updating $\boldsymbol{\theta}_k^\mu$ following \eqref{eq_gd_mu}. Common choices for $W(t)$ include the Ornstein-Uhlenbeck process \cite{uhlenbeck1930theory} or modeling it as a Gaussian distribution. Specifically, let $t_i$ denote the system time at the $i$-th observation. Given system state $\boldsymbol{x}(t_i) = \boldsymbol{x}_{t_i}$, by applying
\begin{equation}\label{eq_exp_policy}
\boldsymbol{\bar u}(t_i) = \boldsymbol{\mu}(\boldsymbol{x}(t_i), \boldsymbol{\theta}_k^\mu) + \sigma(t_i)W(t_i)
\end{equation} into the unknown dynamics in \eqref{eq_unknown_dyn}, the system state at the next time step, $\boldsymbol{x}(t_i+1)$ is observed, where $\sigma(t_i)\geq 0$ is a time-decay function designed to ensure that the effect of $W(t_i)$ on $\boldsymbol{\mu}$ gradually decreases as $\boldsymbol{\theta}_k^\mu$ approaches the optimal value $\boldsymbol{\theta}^{\mu*}$. To differentiate between system state-input variables and observed constant system state-input data pairs, we denote $(\boldsymbol{x}_{t_i}, \boldsymbol{\bar u}_{t_i}, \boldsymbol{x}_{t_i}^+)$ as the observed constant tuple corresponding to $(\boldsymbol{x}(t_i), \boldsymbol{\bar u}(t_i), \boldsymbol{x}(t_i+1))$. For notational brevity, we write $(\boldsymbol{x}_i, \boldsymbol{u}_i, \boldsymbol{x}_i^+)\coloneqq (\boldsymbol{x}_{t_i}, \boldsymbol{\bar u}_{t_i}, \boldsymbol{x}_{t_i}^+)$ throughout the remainder of this paper. The observed tuples $(\boldsymbol{x}_i, \boldsymbol{u}_i, \boldsymbol{x}_i^+)$ are stored in a finite-sized data memory $\mathcal{D}_i$, which has a fixed maximum capacity $N$, and is defined as follows:
\[\mathcal{D}_i = \{(\boldsymbol{x}_1, \boldsymbol{u}_1, \boldsymbol{x}_1^+), (\boldsymbol{x}_2, \boldsymbol{u}_2, \boldsymbol{x}_2^+),\cdots, (\boldsymbol{x}_i, \boldsymbol{u}_i, \boldsymbol{x}_i^+)\}, i\leq N.\] For $i>N$, the earliest $i-N$ observed tuples are discarded to accommodate new data. The value of $N$ is typically chosen to be large. At any iteration $k$, a mini-batch $\mathcal{D}_k$ of $\tilde N$ tuples ($\tilde N\ll N$) is sampled from $\mathcal{D}_i$, with its index set $\mathcal{I}_k$ drawn uniformly from all subsets of $\{1,2,\cdots,i\}$ with size $\tilde N$, provided that $i\geq \tilde N$. This sampled $\mathcal{D}_k$ is then used to construct \eqref{xyudata} and estimate the gradients in \eqref{eq_gd_thetaf}, \eqref{TD_update}, and \eqref{eq_approximate_pg}.

Under the online setting, data collection and parameter updates are performed concurrently. Moreover, the gradients of \eqref{eq_gd_thetaf}, \eqref{TD_update}, and \eqref{eq_approximate_pg} computed over $\mathcal{D}_k$ serve as an unbiased estimation of those obtained from the full dataset $\mathcal{D}_i$.

\begin{figure}[ht]
    \centering
\includegraphics[width=0.85\linewidth]{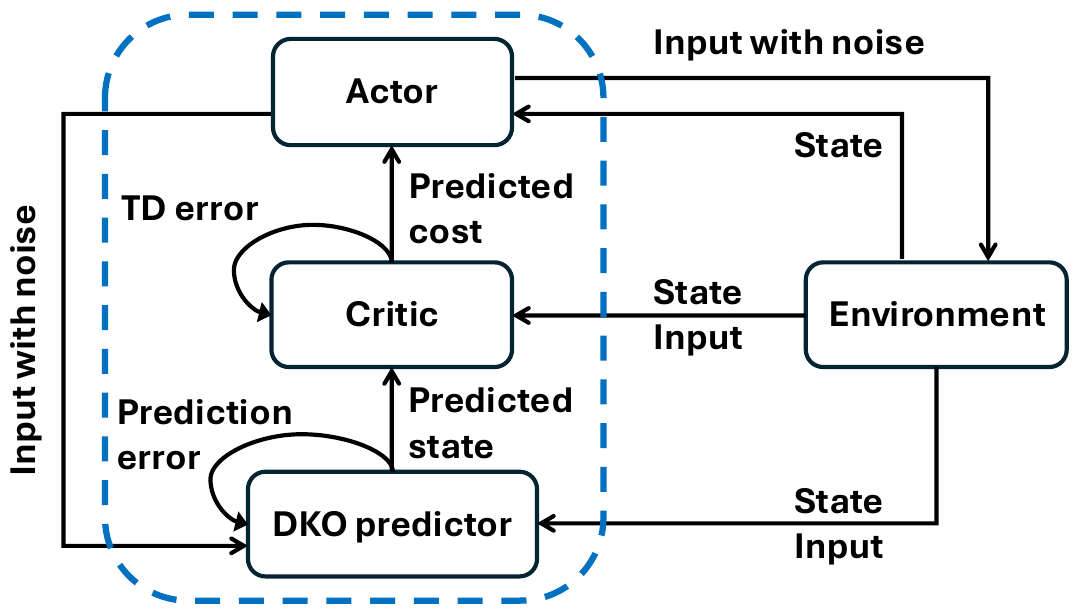}
    \caption{PGDK framework under online implementation.}
    \label{fig:alg_online}
\end{figure}

In summary, Algorithm~\ref{alg2} presents the online variant of PGDK, which implements the PGDK framework by computing gradients over sampled mini-batches rather than a fixed offline dataset. The overall structure of the algorithm is illustrated in Fig.~\ref{fig:alg_online}.
\begin{algorithm}[ht]
\SetAlgoLined
\LinesNumbered
Initialize $\boldsymbol{\theta}_0^f\in\mathbb{R}^p$, $\boldsymbol{\theta}_0^J\in\mathbb{R}^s$, and $\boldsymbol{\theta}_0^\mu\in\mathbb{R}^q$ for $\boldsymbol{g}(\cdot,\boldsymbol{\theta}^f)$, $ V(\cdot,\boldsymbol{\theta}^J)$, and $\boldsymbol{\mu}(\cdot, \boldsymbol{\theta}^\mu)$, respectively.
Set the iteration index $k=0$ and step size sequences $\{\alpha_k^f\}_{k=0}^K$, $\{\alpha_k^J\}_{k=0}^K$, $\{\alpha_k^\mu\}_{k=0}^K$. Specify discount factor $\gamma$, number of episodes $E$, task horizon $T$, batch size $\tilde N$, time-decay function $\sigma(t)$, and data memory.\\
\For{$\textnormal{episode}= 1,2,\cdots, E$}{Reset the initial state $\boldsymbol{x}_0$ and initialize the noise process $W$.\\
    \For{$i=0,1,\cdots,T$}{
       Execute the control input $\boldsymbol{\bar u}(t_i) = \boldsymbol{\mu}(\boldsymbol{x}(t_i), \boldsymbol{\theta}_k^\mu) + \sigma(t_i)W(t_i)$, observe the resulting $\boldsymbol{x}(t_i+1)$, and store the observed tuple $(\boldsymbol{x}_{t_i}, \boldsymbol{\bar u}_{t_i}, \boldsymbol{x}_{t_i+1})$ in the data memory.\\
       Sample mini-batch $\mathcal{D}_k$ of $\tilde N$ tuples uniformly from the data memory.\\
        Update $\boldsymbol{\theta}_k^f$ analogous to \eqref{eq_gd_thetaf} using the batch gradient, where $A_k$, $B_k$, and $C_k$ are computed following \eqref{lmn} using $\mathcal{D}_k$.\\ Update $\boldsymbol{\theta}_k^J$ and $\boldsymbol{\theta}_k^\mu$ in a manner analogous to \eqref{TD_update} and \eqref{eq_gd_L2}, respectively, using the gradient computed over $\mathcal{D}_k$.\\
       Set $k = k+1$.
        }}
\caption{Policy Gradient with Deep Koopman Representation (PGDK) — online implementation}\label{alg2}
\end{algorithm}
\subsection{Numerical Simulations}
In this subsection, we apply the proposed algorithms to derive closed-loop optimal controllers for both a linear time-invariant (LTI) system and a nonlinear inverted pendulum example. Following this, we compare the performance of the proposed methods with related baseline algorithms, highlighting key differences in control strategies and convergence behavior. Finally, we provide a detailed performance analysis, offering insights into the advantages and limitations of the proposed PGDK framework in various settings.
\subsubsection{LTI System}
Consider the following LTI dynamics: \begin{equation}
    \boldsymbol{x}(t+1) = \begin{bmatrix}
        &0.5 &0.5\\& 0 & 1
    \end{bmatrix} \boldsymbol{x}(t) + \begin{bmatrix}
        0\\1
    \end{bmatrix} u(t), \quad \boldsymbol{x}(0) = \boldsymbol{x}_0, \nonumber
\end{equation}
where $[-5,-5]'\leq\boldsymbol{x}(t)\leq [5,5]'$ and $-1\leq u(t)\leq 1$ denote the system state and control input at time $t$, respectively. 

\textbf{Setup.} For each simulation episode, the initial state $\boldsymbol{x}_0$ is uniformly sampled from the interval $[-0.1,-0.1]'$ and $[0.1,0.1]'$, and each episode is terminated either when $t\geq 50$. 
The objective of this example is to design a closed-loop optimal controller for driving the system state toward the goal state $\boldsymbol{x}_{\mathrm{goal}} = [1,1]'$ starting from any given initial state, for which we design the following stage cost function: \begin{equation}
    c(\boldsymbol{x}(t),u(t)) = (\boldsymbol{x}(t)-\boldsymbol{x}_{\mathrm{goal}})'(\boldsymbol{x}(t)-\boldsymbol{x}_{\mathrm{goal}}) + 0.001u(t)^2. \nonumber
\end{equation}
To conduct the simulation, we construct the deep Koopman basis function $\boldsymbol{g}(\cdot,\boldsymbol{\theta}^f):\mathbb{R}^2\rightarrow\mathbb{R}^4$ to lift the original system states into a higher-dimensional space, which is applied in both the offline and online PGDK algorithms. To satisfy Assumption~\ref{asp1}, the functions $\boldsymbol{g}$ and $V$ are implemented as two-layer DNNs with $ReLU$ activation functions, consisting of $400$ and $300$ neurons per layer, respectively. The policy $\boldsymbol{\mu}$ uses a $ReLU$ function with $400$ neurons in the first hidden layer and a 
$tanh$ function with $300$ neurons in the second hidden layer. In particular, we execute the online PGDK algorithm first to generate a data memory, which is then used as the training dataset for the offline PGDK algorithm. All DNNs are trained via the \textit{Adam} optimizer \cite{kingma2014adam}.

\textbf{Evaluation.} In this simulation, we are interested in evaluating how the proposed offline and online PGDK perform on a simple control task compared to the classical optimal control technique, the linear quadratic regulator (LQR) with horizon length $50$, which has access to the exact system dynamics and employs the same stage cost function. For both PGDK and LQR, the system states and control inputs are clipped to satisfy state and input constraints. To assess the algorithm performance, we compare them by presenting both the simulation costs and the optimal trajectories achieved by each method, using identical initial states to ensure a fair comparison. To mitigate the influence of randomness in the DNN training, the entire experiment is repeated over $5$ independent trials.

\begin{figure}[ht]
    \centering
\includegraphics[width=0.9\linewidth]{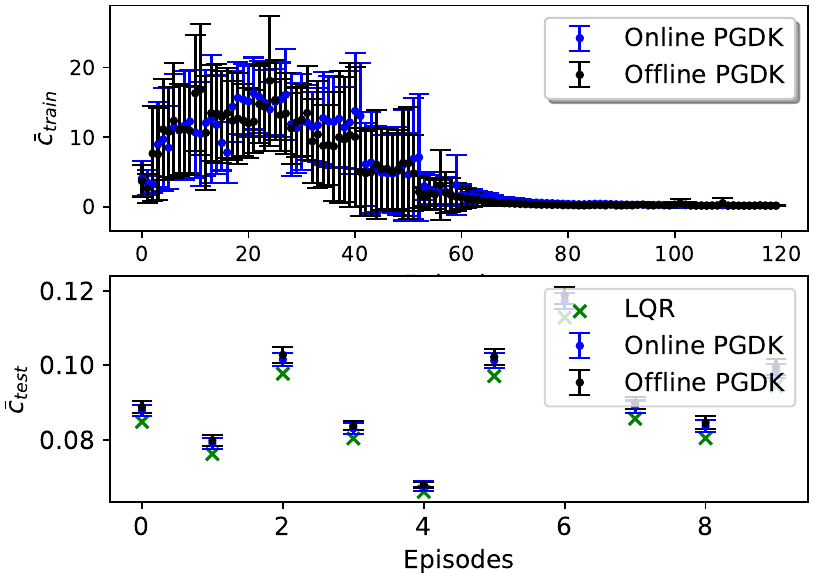}
    \caption{Learning and testing stage cost using the same initial states. Here, $\bar{c}$ denotes the averaged stage cost over each episode to account for the variance of different initial states. The solid line represents the mean stage cost across $5$ experiment trials, while the shaded region and error bars indicate the standard deviation.}
    \label{fig:LTI_cost}
\end{figure}

\begin{figure}[ht]
    \centering
\includegraphics[width=0.9\linewidth]{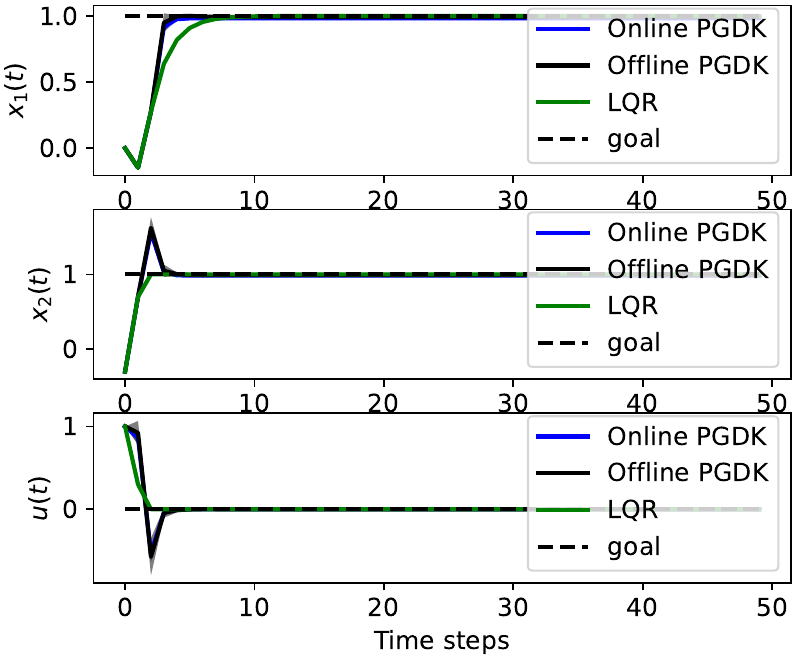}
    \caption{Trajectories from PGDK and LQR.}
    \label{fig:LTI_traj}
\end{figure}

\textbf{Analysis.} As shown in the first subplot of Fig.~\ref{fig:LTI_cost}, the online PGDK algorithm demonstrates a convergence rate close to that of the offline PGDK approach with respect to the number of training episodes. Furthermore, the second subplot of Fig.~\ref{fig:LTI_cost} shows that the closed-loop controller learned via online PGDK attains stage costs that are consistently closer to those of the benchmark LQR controller relative to the offline PGDK implementation, when evaluated from identical initial states. This improvement in stage cost can be attributed to the fact that the offline PGDK is more prone to converging to local minima. In contrast, the online PGDK, which updates gradients using mini-batches sampled from a continually refreshed replay buffer, is more likely to escape local minima. Notably, the error bars in the second subplot indicate that the best-performing instances of the proposed PGDK algorithms across $5$ trials attain performance close to the LQR controller, underscoring the potential of the PGDK framework to achieve control performance close to classical optimal control methods when properly trained.
Fig.~\ref{fig:LTI_traj} illustrates an example trajectory comparison between the proposed PGDK algorithms and the LQR controller. It can be observed that both online and offline PGDK algorithms tend to apply more aggressive control inputs than the LQR controller in this LTI system scenario. This behavior results in faster convergence toward the goal state, albeit at the cost of potentially higher control energy. The LQR controller, on the other hand, follows a smoother trajectory with more conservative control inputs, reflecting its optimality under a quadratic cost formulation.

\subsubsection{Simulated Inverted Pendulum}
We now illustrate the performance of the proposed algorithms using the nonlinear inverted pendulum example, of which the system dynamics are given by \begin{equation}\label{eq_pen_dyn}
    \begin{cases}
    \theta(t+1) = \theta(t) + \dot{\theta}(t+1) \Delta t,\\
      \dot{\theta}(t+1) = \dot{\theta}(t) + (\frac{-3g\sin(\theta(t) + \pi)}{2l} + \frac{3u(t)}{ml^2})\Delta t,
\end{cases}
\end{equation} where $g$, $m$, and $l$ represent the acceleration due to gravity, the mass of the pendulum, and the length of the pendulum, respectively. Here, the system state is defined as $\boldsymbol{x}(t) = [\theta(t), \dot{\theta}(t)]'$ with the lower bound and upper bound of $[-\pi,-8]'$ and $[\pi,8]'$, respectively, and $u(t)$ denotes the continuous scalar torque input, constrained between $-2$ and $2$. 

\textbf{Setup.} For each simulation episode, the initial state $\boldsymbol{x}(0)$ is sampled uniformly from the interval $[-\pi,-1]'$ and $[\pi,1]'$ and an episode terminates if $t>200$. The goal is to balance the pendulum to the upright position, represented by $\boldsymbol{x}_{\mathrm{goal}} = [0, 0]'$. To achieve this, the following stage cost function is defined: $$c(\boldsymbol{x}(t), \boldsymbol{u}(t)) = \theta(t)^2 + 0.1\dot{\theta}(t)^2 + 0.001u(t)^2.$$ To conduct the simulation, we set $\Delta t = 0.02s$, and lift the system states using the deep Koopman basis function $\boldsymbol{g}(\cdot,\boldsymbol{\theta}^f):\mathbb{R}^2\rightarrow\mathbb{R}^8$, which is utilized in both the offline and online PGDK algorithms. The functions $\boldsymbol{g}$ and $V$ adopt the same architectures and activation functions as in the LTI example. The policy $\boldsymbol{\mu}$ is implemented as a two-layer DNN with $ReLU$ functions in the hidden layers ($400$ and $300$ neurons, respectively) and a $tanh$ output layer. The online PGDK algorithm is executed first, and the data memory generated during this process is retained and subsequently used as the training dataset for the offline PGDK implementation. We use the \textit{Adam} optimizer for DNN training.

\textbf{Evaluation.} This simulation evaluates the convergence rate of the proposed online PGDK algorithm (trained for $200$ episodes) compared with existing on-policy RL methods and assesses the optimality of the resulting closed-loop controller relative to model-based optimal control strategies. We compare PGDK with four additional approaches: \begin{itemize}
    \item DDPG: A representative actor-critic RL algorithm trained for $1000$ episodes using the same actor–critic learning rates and optimizer as PGDK.
    \item MPC: Model predictive control with the exact system dynamics, horizon length $50$, and input constraints applied during optimization.
    \item DKMPC-v1: MPC using a learned deep Koopman model trained offline using $9000$ state–input pairs generated by applying control inputs sampled from a normal distribution \cite{dk}, with horizon $8$ and input constraints.
    \item DKMPC-v2: MPC using the dynamics learned by PGDK, horizon $20$, with input constraints.
\end{itemize} For PGDK and DDPG, states and inputs are clipped to satisfy constraints. Policy optimality is evaluated via the trial cost $J = \sum_{t=0}^{200} c(\boldsymbol{x}(t), \boldsymbol{u}(t))$, across all methods. The comparison uses identical stage cost functions and seven initial states, ordered from closest to farthest from the goal: $[\theta_0, \dot{\theta}_0]=[\pi/12, -1]$, $[-\pi/12, -1]$, $[\pi/4, 1]$, $[-\pi/4, 1]$, $[\pi/2, 0]$, $[-\pi/2, 0]$, $[\pi, 0]$.

\begin{figure}[ht]
    \centering
\includegraphics[width=0.9\linewidth]{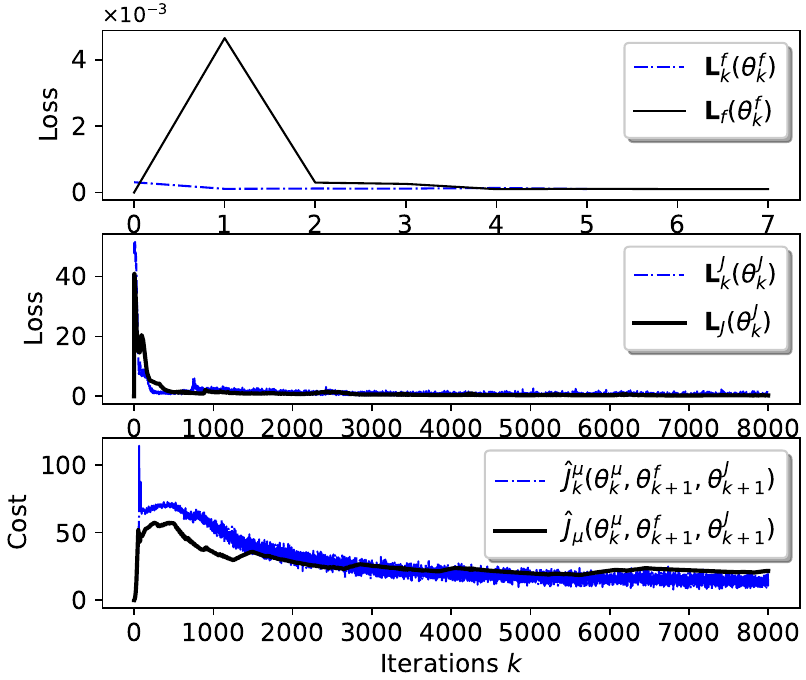}
    \caption{Learning losses of the online (blue) and offline (black) PGDK, where $L_f$, $L_J$, and $\hat{J}$ represent the loss functions computed over the entire dataset at iteration $k$, while $L_k^f$, $L_k^J$, and $\hat{J}_k^\mu$ denote their corresponding values computed using the sampled data batches.}
    \label{fig:loss_his}
\end{figure}

\begin{figure}[ht]
    \centering
\includegraphics[width=0.8\linewidth]{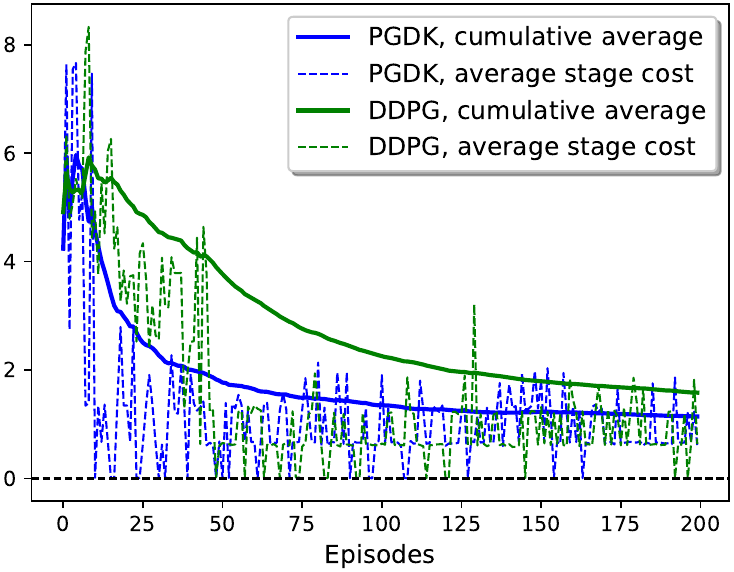}
    \caption{Learning stage cost of online PGDK and DDPG, where the dashed line denotes the average stage cost of each episode to account for the variability in initial states, and the solid lines denote the cumulative average stage cost.}
    \label{fig:reward}
\end{figure}

\begin{figure}[ht]
    \centering
\includegraphics[width=0.92\linewidth]{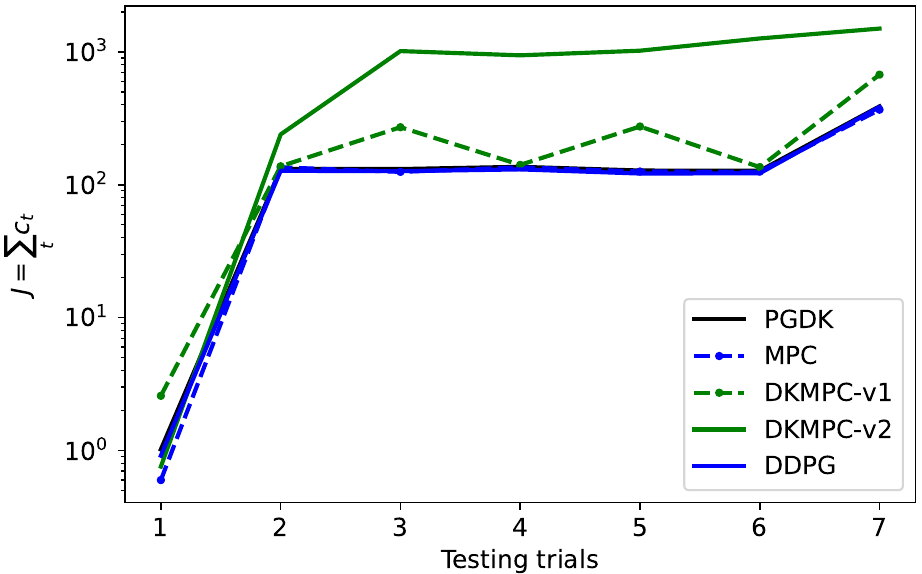}
    \caption{Testing cost plotted on a logarithmic scale.}
    \label{fig:J_pgdkmpc}
\end{figure}

\textbf{Analysis.} 
As demonstrated in Fig.~\ref{fig:loss_his}, both offline and online PGDK methods exhibit asymptotic convergence, consistent with theoretical expectations. Notably, offline PGDK reaches policy convergence in fewer iterations since it computes gradients over the entire dataset, while the online PGDK estimates gradients using sampled data batches. However, in practical applications, offline PGDK typically demands more computational time due to the necessity of processing the full dataset for parameter updates.
As can be seen in Figs. \ref{fig:reward}-\ref{fig:J_pgdkmpc}, for the inverted pendulum example, the proposed online PGDK can reach DDPG's performance but with better data efficiency since it requires fewer episodes ($10$ episodes) to find the optimal policy compared to the DDPG ($60$ episodes).
Fig.~\ref{fig:J_pgdkmpc} further compares online PGDK with model-based MPC under identical initial states. The testing trial costs show that both online PGDK and DDPG approach the performance of MPC with access to exact system dynamics. Notably, online PGDK outperforms MPC when the latter relies on trajectory propagation via deep Koopman models (both DKMPC-v1 and DKMPC-v2), which are prone to cumulative model approximation errors.
\section{Concluding Remarks}\label{Conc}
In this paper, we have described a data-driven policy gradient approximation framework, termed policy gradient with deep Koopman representation (PGDK), which is designed to derive a closed-loop optimal controller for systems with unknown dynamics. Additionally, we have introduced both an offline and an online algorithm to implement the proposed framework. The key contribution of PGDK lies in its integration of deep Koopman learning within the concepts of actor-critic methods to approximate the policy gradient in \eqref{eq_pg} with \eqref{eq_approximate_pg}, enabling the simultaneous approximation of system dynamics, cost functions, and optimal policies. This integration enhances data efficiency and accelerates convergence compared to conventional deterministic policy gradient methods. Furthermore, we present a theoretical convergence analysis of the proposed framework, supported by numerical simulations. Finally, through experiments on an LTI system and a pendulum example, we demonstrate that the closed-loop optimal controller derived via the proposed algorithms for an unknown dynamical system achieves performance close to that of an optimal controller designed with known system dynamics within finite tuning iterations. 


\section{Appendix}
In this section, we present the proofs of the theoretical results discussed in Section \ref{analysis}.
\subsection{Proof of Lemma~\ref{thm1}}
In this subsection, we present the convergence proof for the proposed updating rule in \eqref{eq_gd_thetaf}. To proceed, we first recall the loss function $\bL_f(A,B,C,\boldsymbol{\theta}^f)$ from \eqref{eq_L_f} and data matrices from \eqref{xyudata}. Given that the DNN $\boldsymbol{g}(\boldsymbol{x},\boldsymbol{\theta}^f)$ is constructed using a linear function approximator of the form: $\boldsymbol{g}(\boldsymbol{x},\boldsymbol{\theta}^f) = \boldsymbol{\phi}_f(\boldsymbol{x})\boldsymbol{\theta}^f$, where $\parallel\boldsymbol{\phi}_f(\boldsymbol{x}) \parallel\leq 1$, we reformulate $\bL_f(A,B,C,\boldsymbol{\theta}^f)$ into a compact form using the definition of the Frobenius norm, given by \begin{equation}
    \begin{aligned}
        &\bL_f(A, B, C, \boldsymbol{\theta}^f)\\=& \frac{1}{2N}(\parallel \boldsymbol{\bar\Phi}_f\cdot\boldsymbol{\theta}^f\!\!-\!\!A\boldsymbol{\Phi}_f\cdot\boldsymbol{\theta}^f \!\!-\!\!B\mathbf{U}\parallel_F^2 + \parallel \mathbf{\bar X} \!- \!C\boldsymbol{\bar\Phi}_f\cdot\boldsymbol{\theta}^f\parallel_F^2), \nonumber
    \end{aligned}
\end{equation}
where $\boldsymbol{\bar\Phi}_{f} = [\boldsymbol{\phi}_f(\boldsymbol{x}_1^+),\boldsymbol{\phi}_f(\boldsymbol{x}_2^+),\cdots,\boldsymbol{\phi}_f(\boldsymbol{x}_{N}^+)]\in\mathbb{R}^{r\times Np}$ and $\boldsymbol{\Phi}_f = [\boldsymbol{\phi}_f(\boldsymbol{x}_1),\boldsymbol{\phi}_f(\boldsymbol{x}_2),\cdots,\boldsymbol{\phi}_f(\boldsymbol{x}_{N})]\in\mathbb{R}^{r\times Np}$ are constant matrices. Here, we denote $\boldsymbol{\Phi}_f\cdot\boldsymbol{\theta}^f=[\boldsymbol{\phi}_f(\boldsymbol{x}_1)\boldsymbol{\theta}^f,\boldsymbol{\phi}_f(\boldsymbol{x}_2)\boldsymbol{\theta}^f,\cdots,\boldsymbol{\phi}_f(\boldsymbol{x}_{N})\boldsymbol{\theta}^f]\in\mathbb{R}^{r\times N}$ for the sake of notational simplicity. Note that $\boldsymbol{\Phi}_f\cdot(\boldsymbol{\theta}_1^f + \boldsymbol{\theta}_2^f) = \boldsymbol{\Phi}_f\cdot\boldsymbol{\theta}_1^f + \boldsymbol{\Phi}_f\cdot\boldsymbol{\theta}_2^f$, $\parallel\boldsymbol{\Phi}_f\cdot\boldsymbol{\theta}^f\parallel\leq \parallel \boldsymbol{\Phi}_f\parallel\parallel\boldsymbol{\theta}^f\parallel$, and $\nabla_{\boldsymbol{\theta}^f}\boldsymbol{\Phi}_f\cdot\boldsymbol{\theta}^f = \boldsymbol{\Phi}_f$.

For brevity, we denote $\Delta\boldsymbol{\theta}_k^f = \boldsymbol{\theta}_k^f-\boldsymbol{\theta}^{f*}$ and $\nabla_{\boldsymbol{\theta}^f}\bL_k^f = \nabla_{\boldsymbol{\theta}^f}\bL_f(A_k, B_k, C_k, \boldsymbol{\theta}_k^f)$. To establish that $\lim_{k\rightarrow\infty}\parallel\Delta\boldsymbol{\theta}_k^f\parallel^2=0$, we begin by taking the squared norm after subtracting $\boldsymbol{\theta}^{f*}$ on both sides of \eqref{eq_gd_thetaf}, which results in:
\begin{equation}\label{eq_thm1_pf_1}
    \begin{aligned}
&\parallel\boldsymbol{\theta}_{k+1}^f-\boldsymbol{\theta}^{f*}\parallel^2 \\=&\parallel\boldsymbol{\theta}_k^f-\boldsymbol{\theta}^{f*} - \alpha_k^f\nabla_{\boldsymbol{\theta}^f}\bL_k^f\parallel^2 \\=& \parallel \Delta\boldsymbol{\theta}_k^f \parallel^2\!\! - 2\alpha_k^f \langle \nabla_{\boldsymbol{\theta}^f}\bL_k^f, \Delta\boldsymbol{\theta}_k^f \rangle +\!\!\parallel\alpha_k^f\nabla_{\boldsymbol{\theta}^f}\bL_k^f\parallel^2\!\!\!.\end{aligned}
\end{equation}
The convergence proof of \eqref{eq_thm1_pf_1} consists of two main parts. First, we establish an upper bound for $-\langle \nabla_{\boldsymbol{\theta}^f}\bL_k^f, \Delta\boldsymbol{\theta}_k^f \rangle$. Then, we derive an upper bound for $\parallel\alpha_k^f\nabla_{\boldsymbol{\theta}^f}\bL_k^f\parallel^2$.

To this end, we derive a upper bound of $-\langle \nabla_{\boldsymbol{\theta}^f}\bL_k^f, \Delta\boldsymbol{\theta}_k^f \rangle$ by expanding $\nabla_{\boldsymbol{\theta}^f}\bL_k^f$ and using $\nabla_{\boldsymbol{\theta}^f}\bL_f(A^*, B^*, C^*, \boldsymbol{\theta}^{f*})=0$, yielding:
\begin{equation}\label{eq_thm1_pf_2}
\begin{aligned}
&\langle\nabla_{\boldsymbol{\theta}^f}\bL_k^f(A_k, B_k, C_k, \boldsymbol{\theta}_k^f), \Delta\boldsymbol{\theta}_k^f \rangle\\ =& \langle\nabla_{\boldsymbol{\theta}^f}\bL_f(A_k, B_k, C_k, \boldsymbol{\theta}_k^f) - \nabla_{\boldsymbol{\theta}^f}\bL_f(A_k, B_k, C_k, \boldsymbol{\theta}^{f*}) \\&+ \nabla_{\boldsymbol{\theta}^f}\bL_f(A_k, B_k, C_k, \boldsymbol{\theta}^{f*}), \Delta\boldsymbol{\theta}_k^f \rangle\\=& \frac{1}{N}(\parallel \boldsymbol{\bar\Phi}_f-A_k\boldsymbol{\Phi}_f\parallel_F^2 + \parallel C_k\boldsymbol{\bar\Phi}_f\parallel_F^2)\parallel\Delta\boldsymbol{\theta}_k^f\parallel^2\\&+\frac{1}{2}\Big(\bL_f(A_k, B_k, C_k, \boldsymbol{\theta}^{f*}) - \bL_f(A_k, B_k, C_k, \boldsymbol{\theta}_k^f) \\& - \frac{1}{N}(\parallel (\boldsymbol{\bar\Phi}_f-A_k\boldsymbol{\Phi}_f)\cdot\Delta\boldsymbol{\theta}_k^f\parallel_F^2 + \parallel C_k\boldsymbol{\bar\Phi}_f\cdot\Delta\boldsymbol{\theta}_k^f\parallel_F^2 )\Big)\\=& \frac{3}{2N}(\parallel \boldsymbol{\bar\Phi}_f-A_k\boldsymbol{\Phi}_f\parallel_F^2 + \parallel C_k\boldsymbol{\bar\Phi}_f\parallel_F^2)\parallel\Delta\boldsymbol{\theta}_k^f\parallel^2\\& - \frac{1}{N}(\parallel (\boldsymbol{\bar\Phi}_f - A_k\boldsymbol{\Phi}_f )\cdot\Delta\boldsymbol{\theta}_k^f\parallel_F^2 + \parallel C_k\boldsymbol{\bar\Phi}_f\cdot\Delta\boldsymbol{\theta}_k^f\parallel_F^2 ), 
\end{aligned}
\end{equation}
where the last equality of \eqref{eq_thm1_pf_2} is obtained by applying the Taylor expansion.  Furthermore, based on \eqref{eq_thm1_pf_2}, we obtain the following result: 
\begin{equation}\label{eq_thm1_pf_3}
    \begin{aligned}
        &-\langle \nabla_{\boldsymbol{\theta}^f}\bL_k^f, \Delta\boldsymbol{\theta}_k^f \rangle\\ \leq & -\frac{1}{2N}(\parallel \boldsymbol{\bar\Phi}_f-A_k\boldsymbol{\Phi}_f\parallel_F^2 + \parallel C_k\boldsymbol{\bar\Phi}_f\parallel_F^2)\parallel\Delta\boldsymbol{\theta}_k^f\parallel^2.
    \end{aligned}
\end{equation}
To derive an upper bound for \eqref{eq_thm1_pf_3}, we utilize the cyclic property of the trace operator and apply the Frobenius norm bound and the matrix Cauchy–Schwarz inequality, which yields
\begin{equation}\label{eq_thm1_Lf1}
\begin{aligned}
\small
&\frac{1}{2N}(\parallel \boldsymbol{\bar\Phi}_f\parallel_F^2 \!+\! \parallel A_k\boldsymbol{\Phi}_f\parallel_F^2 \!+\! \parallel C_k\boldsymbol{\bar\Phi}_f\parallel_F^2 \!\!- 2\langle \boldsymbol{\bar\Phi}_f,\! A_k\boldsymbol{\Phi}_f\rangle_F)\\ \geq &\frac{1}{2N}(\omega_f(Np +\!\parallel A_k\parallel_F^2\!+\!\parallel C_k\parallel_F^2)\!-\!2\sqrt{N} \parallel A_k\parallel_F)\eqqcolon L_{f1},
\end{aligned}
\end{equation}
where $\omega_f= \min\{\lambda_{\mathrm{min}}(\boldsymbol{\bar\Phi}_f\boldsymbol{\bar\Phi}_f'), \lambda_{\mathrm{min}}(\boldsymbol{\Phi}_f\boldsymbol{\Phi}_f')\}$.

To establish an upper bound for  $\parallel\nabla_{\boldsymbol{\theta}^f}\bL_k^f\parallel^2$, we proceed by utilizing the definition of $\bL_f$, leading to the following result: 
 \begin{equation}
\begin{aligned}
&\parallel\nabla_{\boldsymbol{\theta}^f}\bL_k^f\parallel^2 
\\ =&\frac{1}{N} \parallel(\boldsymbol{\bar\Phi}_f-A_k\boldsymbol{\Phi}_f)'(\boldsymbol{\bar\Phi}_f\cdot\boldsymbol{\theta}_k^f-A\boldsymbol{\Phi}_f\cdot\boldsymbol{\theta}_k^f - B_k\mathbf{U})\\&-( C_k\boldsymbol{\bar\Phi}_f)'(\mathbf{\bar X} - C\boldsymbol{\bar\Phi}_f\cdot\boldsymbol{\theta}_k^f)\parallel_F^2
\\ \leq& \frac{2}{N}(\parallel\boldsymbol{\bar\Phi}_f-A_k\boldsymbol{\Phi}_f\parallel_F^2\parallel\boldsymbol{\bar\Phi}_f\cdot\boldsymbol{\theta}_k^f-A_k\boldsymbol{\Phi}_f\cdot\boldsymbol{\theta}_k^f - B_k\mathbf{U}\parallel_F^2\\& +\parallel C_k\boldsymbol{\bar\Phi}_f\parallel_F^2\parallel\bar{\bX} - C_k\boldsymbol{\bar\Phi}_f\cdot\boldsymbol{\theta}_k^f\parallel_F^2). \nonumber
\end{aligned}
\end{equation} Given that $[A_k\ B_k]$ and $C_k$ are obtained via the least squares solutions in \eqref{lmn}, the system states and control inputs are bounded, and $\boldsymbol{g}$ is Lipschitz continuous, the norms of the matrices $A_k$, $B_k$, and $C_k$ \eqref{lmn} remain bounded. We denote these bounds as: $\parallel A_k\parallel\leq c_1$, $\parallel B_k\parallel\leq c_2$, and $\parallel C_k\parallel\leq c_3$ throughout this proof. Furthermore, let the residuals from the least squares solutions be bounded as \cite{van1991total}: $\parallel\boldsymbol{\bar\Phi}_f\cdot\boldsymbol{\theta}_k^f-A_k\boldsymbol{\Phi}_f\cdot\boldsymbol{\theta}_k^f - B_k\mathbf{U}\parallel_F^2\leq W_1$, $\parallel\bar{\bX} - C_k\boldsymbol{\bar\Phi}_f\cdot\boldsymbol{\theta}_k^f\parallel_F^2\leq W_2$, which leads to following upper bound:  
\begin{equation}\label{eq_thm1_Lf2}
\parallel\nabla_{\boldsymbol{\theta}^f}\bL_k^f\parallel^2\leq (2(1+c_1^2)W_1 + 2c_3^2 W_2)/N \coloneqq L_{f2}.
\end{equation}
Finally, applying the derived bounds $L_{f1}$ from \eqref{eq_thm1_Lf1} and $L_{f2}$ from \eqref{eq_thm1_Lf2} to \eqref{eq_thm1_pf_1}, and using $1-x\leq e^{-x}$ for real $x$, yields:
\begin{equation}\label{eq_thm1_pf_4}
\small
    \begin{aligned}
&\parallel\Delta\boldsymbol{\theta}_{k+1}^f\parallel^2 \\ \leq &(1-2\alpha_k^f L_{f1})\parallel \Delta\boldsymbol{\theta}_k^f \parallel^2 + (\alpha_k^f)^2 L_{f2}\\ \leq &\prod_{i=0}^{k} (1-2\alpha_i^f L_{f1})\parallel \Delta\boldsymbol{\theta}_0^f \parallel^2 +  L_{f2}\sum_{j=0}^k (\alpha_j^f)^2\!\!\!\prod_{i=j+1}^{k} (1-2\alpha_i^f L_{f1})\\ \leq & \parallel \Delta\boldsymbol{\theta}_0^f \parallel^2e^{-2 L_{f1}\sum_{i=0}^k\alpha_i^f} +  L_{f2}\sum_{j=0}^k (\alpha_j^f)^2\!\!\!\prod_{i=j+1}^{k} (1-2\alpha_i^f L_{f1}).\end{aligned}
\end{equation}
Hence, to achieve $\lim_{k\rightarrow\infty}\parallel\Delta\boldsymbol{\theta}_k^f\parallel^2=0$, one needs to choose step size satisfying $\sum_{k=0}^\infty\alpha_k^f=\infty$ and $\sum_{k=0}^\infty(\alpha_k^f)^2<\infty$. 

It can be observed that, by selecting a decaying step size for \eqref{eq_thm1_cor}, specifically, $\alpha_k^f = \frac{\beta_f}{2+k}$, where $\beta_f = \frac{1}{L_{f1}}$, the following holds: \begin{equation}
    \parallel\boldsymbol{\theta}_k^f-\boldsymbol{\theta}^{f*}\parallel^2\leq \frac{\nu_f}{2+k},
\end{equation}
where $\nu_f = \max\{\beta_f^2L_{f2},\quad  2\parallel\boldsymbol{\theta}_0^f-\boldsymbol{\theta}^{f*}\parallel^2\}$. The main idea to prove this is to show $\parallel\boldsymbol{\theta}_{k+1}^f-\boldsymbol{\theta}^{f*}\parallel^2 \leq \frac{\nu_f}{2 + k + 1}$ using induction.
We start from the case when $k=0$, where one has \[ \parallel\boldsymbol{\theta}_0^f-\boldsymbol{\theta}^{f*}\parallel^2 \leq \frac{\nu_f}{2}.\] Next, if $k\geq 1$, one has: \begin{equation}\label{eq_thm1_cor2}
\begin{aligned}
    &\parallel\boldsymbol{\theta}_{k+1}^f-\boldsymbol{\theta}^{f*}\parallel^2 \\ \leq &(1-2\alpha_k^f L_{f1})\parallel \boldsymbol{\theta}_k^f - \boldsymbol{\theta}^{f*} \parallel^2 + (\alpha_k^f)^2 L_{f2}\\ =& (1-\frac{2\beta_f L_{f1}}{2+k})\parallel \boldsymbol{\theta}_k^f - \boldsymbol{\theta}^{f*} \parallel^2 + \frac{\beta_f^2L_{f2}}{(2+k)^2}  \\ \leq &\frac{(2+k-1)\nu_f}{(2+k)^2} + \frac{(1-2\beta_f L_{f1})\nu_f+\beta_f^2L_{f2}}{(2+k)^2} \\ = &\frac{(2+k-1)\nu_f}{(2+k)^2} + \frac{\beta_f^2L_{f2} - \nu_f}{(2+k)^2} \quad  (\because \beta_f = \frac{1}{L_{f1}})\\ \leq &\frac{\nu_f}{2+k + 1} \quad  (\because (x+1)^2\geq x(x+2)).
    \end{aligned}
\end{equation} 
Moreover, applying constant $0 < \alpha_f < \frac{1}{2L_{f1}}$ to \eqref{eq_thm1_pf_4} and using the geometric series formula ($\sum_{k=0}^\infty ar^k = \frac{a}{1-r}, |r|<1$), we obtain: \begin{equation}\label{eq_thm1_cor}
\begin{aligned}
    &\parallel\Delta\boldsymbol{\theta}_k^f\parallel^2 \\ \leq &(1-2\alpha_f L_{f1})^k \parallel\Delta\boldsymbol{\theta}_0^f\parallel^2 + \alpha_f^2 L_{f2}\sum_{s=0}^{\infty}(1 - 2\alpha_f L_{f1})^s \\ \leq &(1- 2\alpha_f L_{f1})^k \parallel\Delta\boldsymbol{\theta}_0^f\parallel^2 + \frac{\alpha_f L_{f2}}{2L_{f1}}.
    \end{aligned}
\end{equation}
Here, \eqref{eq_thm1_cor} cannot converge to zero if $L_{f2}\neq 0$. $\hfill \blacksquare$

\subsection{Proof of Theorem~\ref{thm2}}
Recall $\hat{J} = \frac{1}{N}\sum_{t^*\in\mathcal{I}_D}\hat{J}_{t^*}$ from \eqref{eq_gd_L2} as the averaged $\hat{J}_{t^*}$ in \eqref{eq_J_hat} computed over $\mathcal{D}$. The goal of this proof is to analyze the convergence of $\nabla_{\boldsymbol{\theta}^\mu}\hat{J}$ while tuning $\boldsymbol{\theta}^f$, $\boldsymbol{\theta}^J$, and $\boldsymbol{\theta}^\mu$ using the proposed PGDK framework.

\subsubsection{Derivation of Lipschitz Constants}
To begin, we need to ensure the approximated policy gradient $\nabla_{\boldsymbol{\theta}^\mu}\hat{J}_{t^*}$ from \eqref{eq_approximate_pg} is Lipschitz continuous. For simplicity and ease of notation, we denote $\boldsymbol{u}_{t^*}^i \coloneqq \mu(\boldsymbol{x}_{t^*}, \boldsymbol{\theta}_i^\mu)$ and $\boldsymbol{\hat{x}}_{t^*+1}^i \coloneqq C_k(A_k\boldsymbol{g}(\boldsymbol{x}_{t^*}, \boldsymbol{\theta}_{k+1}^f) + B_k\boldsymbol{u}_{t^*}^i)$. By following the definition of $\hat{J}_{t^*}$, for any $\boldsymbol{\theta}_1^\mu, \boldsymbol{\theta}_2^\mu\in\mathbb{R}^q$, the following holds:
\begin{equation}\label{eq_lemma1pf_1}
\begin{aligned}
& \parallel \nabla_{\boldsymbol{\theta}^\mu} \hat{J}_{t^*}(\boldsymbol{\theta}_1^\mu) - \nabla_{\boldsymbol{\theta}^\mu} \hat{J}_{t^*}(\boldsymbol{\theta}_2^\mu) \parallel \\ =& \parallel \nabla_{\boldsymbol{u}}c(\boldsymbol{u}_{t^*}^1) \nabla_{\boldsymbol{\theta}^\mu}\boldsymbol{u}_{t^*}(\boldsymbol{\theta}_1^\mu) + \gamma\nabla_{\boldsymbol{\hat x}} V(\boldsymbol{\hat{x}}_{t^*+1}^1)  \nabla_{\boldsymbol{u}} \boldsymbol{\hat{x}}_{t^*+1}^1 \\ & \times \nabla_{\boldsymbol{\theta}^\mu}\boldsymbol{u}_{t^*}(\boldsymbol{\theta}_1^\mu) - \nabla_{\boldsymbol{u}}c(\boldsymbol{u}_{t^*}^2) \nabla_{\boldsymbol{\theta}^\mu}\boldsymbol{u}_{t^*}(\boldsymbol{\theta}_2^\mu) - \gamma\nabla_{\boldsymbol{\hat x}} V(\boldsymbol{\hat{x}}_{t^*+1}^2) \\ &\times \nabla_{\boldsymbol{u}} \boldsymbol{\hat{x}}_{t^*+1}^2 \nabla_{\boldsymbol{\theta}^\mu}\boldsymbol{u}_{t^*}(\boldsymbol{\theta}_2^\mu)\parallel \\
 = & \parallel \nabla_{\boldsymbol{u}}c(\boldsymbol{u}_{t^*}^1) (\nabla_{\boldsymbol{\theta}^\mu}\boldsymbol{u}_{t^*}(\boldsymbol{\theta}_1^\mu) - \nabla_{\boldsymbol{\theta}^\mu}\boldsymbol{u}_{t^*}(\boldsymbol{\theta}_2^\mu))  \\ & + (\nabla_{\boldsymbol{u}}c(\boldsymbol{u}_{t^*}^1) - \nabla_{\boldsymbol{u}}c(\boldsymbol{u}_{t^*}^2)) \nabla_{\boldsymbol{\theta}^\mu}\boldsymbol{u}_{t^*}(\boldsymbol{\theta}_2^\mu) + \gamma\nabla_{\boldsymbol{\hat x}} V(\boldsymbol{\hat{x}}_{t^*+1}^1)  \\ &\times( \nabla_{\boldsymbol{u}} \boldsymbol{\hat{x}}_{t^*+1}^1  \nabla_{\boldsymbol{\theta}^\mu}\boldsymbol{u}_{t^*}(\boldsymbol{\theta}_1^\mu) - \nabla_{\boldsymbol{u}} \boldsymbol{\hat{x}}_{t^*+1}^2 \nabla_{\boldsymbol{\theta}^\mu}\boldsymbol{u}_{t^*}(\boldsymbol{\theta}_2^\mu))  \\ &+ \gamma(\nabla_{\boldsymbol{\hat x}} V(\boldsymbol{\hat{x}}_{t^*+1}^1) - \nabla_{\boldsymbol{\hat x}} V(\boldsymbol{\hat{x}}_{t^*+1}^2))\nabla_{\boldsymbol{u}} \boldsymbol{\hat{x}}_{t^*+1}^2 \nabla_{\boldsymbol{\theta}^\mu}\boldsymbol{u}_{t^*}(\boldsymbol{\theta}_2^\mu)\parallel.
 \end{aligned}
 \end{equation}
Then, by following the norm triangle inequality, subordinance, and submultiplicativity, \eqref{eq_lemma1pf_1} becomes:
\begin{equation}
\begin{aligned}
& \parallel \nabla_{\boldsymbol{\theta}^\mu} \hat{J}_{t^*}(\boldsymbol{\theta}_1^\mu) - \nabla_{\boldsymbol{\theta}^\mu} \hat{J}_{t^*}(\boldsymbol{\theta}_2^\mu) \parallel \\
 \leq & \parallel \nabla_{\boldsymbol{u}}c(\boldsymbol{u}_{t^*}^1)\parallel\parallel \nabla_{\boldsymbol{\theta}^\mu}\boldsymbol{u}_{t^*}(\boldsymbol{\theta}_1^\mu) - \nabla_{\boldsymbol{\theta}^\mu}\boldsymbol{u}_{t^*}(\boldsymbol{\theta}_2^\mu)\parallel \\ &+ \parallel\nabla_{\boldsymbol{u}}c(\boldsymbol{u}_{t^*}^1) - \nabla_{\boldsymbol{u}}c(\boldsymbol{u}_k^2) \parallel \parallel\nabla_{\boldsymbol{\theta}^\mu}\boldsymbol{u}_{t^*}(\boldsymbol{\theta}_2^\mu)\parallel \\ & + \gamma\parallel\nabla_{\boldsymbol{\hat x}} V(\boldsymbol{\hat{x}}_{t^*+1}^1)\parallel \parallel\nabla_{\boldsymbol{u}} \boldsymbol{\hat{x}}_{t^*+1}^1 \nabla_{\boldsymbol{\theta}^\mu}\boldsymbol{u}_{t^*}(\boldsymbol{\theta}_1^\mu) - \nabla_{\boldsymbol{u}} \boldsymbol{\hat{x}}_{t^*+1}^2 \\ &\times \nabla_{\boldsymbol{\theta}^\mu}\boldsymbol{u}_{t^*}(\boldsymbol{\theta}_2^\mu)\parallel + \gamma\parallel \nabla_{\boldsymbol{\hat x}} V(\boldsymbol{\hat{x}}_{t^*+1}^1) - \nabla_{\boldsymbol{\hat x}} V(\boldsymbol{\hat{x}}_{t^*+1}^2)\parallel \\ &\times \parallel \nabla_{\boldsymbol{u}} \boldsymbol{\hat{x}}_{t^*+1}^2 \nabla_{\boldsymbol{\theta}^\mu}\boldsymbol{u}_{t^*}(\boldsymbol{\theta}_2^\mu) \parallel. \nonumber
\end{aligned}
\end{equation}
Finally, given that $\nabla_{\boldsymbol{u}} \boldsymbol{\hat{x}}_{t^*+1}^1 = \nabla_{\boldsymbol{u}} \boldsymbol{\hat{x}}_{t^*+1}^2 = C_kB_k$, and by applying the matrix norm bound and Lipschitz continuity, the following result is derived:
 \begin{equation}
\begin{aligned}
& \parallel \nabla_{\boldsymbol{\theta}^\mu} \hat{J}_{t^*}(\boldsymbol{\theta}_1^\mu) - \nabla_{\boldsymbol{\theta}^\mu} \hat{J}_{t^*}(\boldsymbol{\theta}_2^\mu) \parallel \\
 \leq & L_{c\mu} L_{\mu\theta\theta}\parallel \boldsymbol{\theta}_1^\mu - \boldsymbol{\theta}_2^\mu\parallel + L_{c\mu\mu} L_{\mu\theta}^2\parallel \boldsymbol{\theta}_1^\mu - \boldsymbol{\theta}_2^\mu\parallel   + \gamma L_{vx} L_{\mu\theta\theta} \\ & \times c_2 c_3 \parallel \boldsymbol{\theta}_1^\mu - \boldsymbol{\theta}_2^\mu\parallel  + \gamma L_{vxx}L_{\mu\theta}^2 (c_2 c_3)^2 \parallel \boldsymbol{\theta}_1^\mu - \boldsymbol{\theta}_2^\mu\parallel\\
 = & (L_{c\mu} L_{\mu\theta\theta} + L_{c\mu\mu} L_{\mu\theta}^2 + \gamma L_{vx} L_{\mu\theta\theta}c_2 c_3 + \gamma L_{vxx}L_{\mu\theta}^2(c_2 c_3)^2 ) \\ &\parallel \boldsymbol{\theta}_1^\mu - \boldsymbol{\theta}_2^\mu\parallel \\ \eqqcolon &L_{J\theta\theta}\parallel \boldsymbol{\theta}_1^\mu - \boldsymbol{\theta}_2^\mu\parallel, \nonumber
\end{aligned}
\end{equation} where $L_{c\mu}$, $L_{c\mu\mu}$ is the Lipschitz constants of the stage cost $c$ and $\nabla c$ with regarding $\boldsymbol{\mu}$, respectively. Similarly, $L_{\mu\theta}$ and $L_{\mu\theta\theta}$ represent the Lipschitz constants of $\boldsymbol{\mu}$ and $\nabla\boldsymbol{\mu}$ with regard to $\boldsymbol{\theta}^\mu$, respectively. Moreover, $L_{vx}$ and $L_{vxx}$ denote the Lipschitz constants of $V$ and $\nabla V$ with regarding $\boldsymbol{x}$, respectively. Similarly, one has $$\parallel \nabla_{\boldsymbol{\theta}^\mu}\hat{J}_{t^*}\parallel \leq (L_{c\mu}L_{\mu\theta} +\gamma L_{vx}L_{\mu\theta}c_2 c_3) \eqqcolon L_{J\theta}.$$

\subsubsection{Convergence Analysis of the Estimated Policy Gradient} We now begin the analysis by employing the Taylor expansion for  $\hat{J}(\boldsymbol{\theta}_{k+1}^f, \boldsymbol{\theta}_{k+1}^J,\boldsymbol{\theta}_k^\mu)$ while disregarding higher-order terms, which will lead to the following results:
\begin{equation}
    \begin{aligned}
   &\hat{J}(\boldsymbol{\theta}_{k+1}^f, \boldsymbol{\theta}_{k+1}^J, \boldsymbol{\theta}_{k+1}^\mu) \\=& \hat{J}(\boldsymbol{\theta}_{k+1}^f, \boldsymbol{\theta}_{k+1}^J, \boldsymbol{\theta}_k^\mu) + \langle \nabla_{\boldsymbol{\theta}^\mu}\hat{J}( \boldsymbol{\theta}_{k+1}^f, \boldsymbol{\theta}_{k+1}^J, \boldsymbol{\theta}_k^\mu), \boldsymbol{\theta}_{k+1}^\mu - \boldsymbol{\theta}_k^\mu\rangle \\ &+ \frac{1}{2}\nabla_{\boldsymbol{\theta}^\mu \boldsymbol{\theta}^\mu}\hat{J}(\boldsymbol{\theta}_{k+1}^f, \boldsymbol{\theta}_{k+1}^J, \boldsymbol{\theta}_k^\mu) \parallel \boldsymbol{\theta}_{k+1}^\mu - \boldsymbol{\theta}_k^\mu \parallel^2\\=&  - \alpha_k^\mu \nabla_{\boldsymbol{\theta}^\mu}\hat{J}( \boldsymbol{\theta}_{k+1}^f, \boldsymbol{\theta}_{k+1}^J, \boldsymbol{\theta}_k^\mu)'  \nabla_{\boldsymbol{\theta}^\mu}\hat{J}( \boldsymbol{\theta}_{k+1}^f, \boldsymbol{\theta}_{k+1}^J, \boldsymbol{\theta}_k^\mu) \\ & + \hat{J}(\boldsymbol{\theta}_{k+1}^f, \boldsymbol{\theta}_{k+1}^J, \boldsymbol{\theta}_k^\mu) + \frac{(\alpha_k^\mu)^2}{2}\nabla_{\boldsymbol{\theta}^\mu \boldsymbol{\theta}^\mu}\hat{J}(\boldsymbol{\theta}_{k+1}^f, \boldsymbol{\theta}_{k+1}^J, \boldsymbol{\theta}_k^\mu) \\& \times \parallel \nabla_{\boldsymbol{\theta}^\mu}\hat{J}( \boldsymbol{\theta}_{k+1}^f, \boldsymbol{\theta}_{k+1}^J, \boldsymbol{\theta}_k^\mu)\parallel^2.
    \end{aligned}
\end{equation} 
Since the approximated policy gradient is Lipschitz continuous and rearranging the results yields the following expression:
\begin{equation}\label{eq_thm2_2}
\begin{aligned} &\alpha_k^\mu \parallel\nabla_{\boldsymbol{\theta}^\mu}\hat{J}(\boldsymbol{\theta}_{k+1}^f, \boldsymbol{\theta}_{k+1}^J, \boldsymbol{\theta}_k^\mu) \parallel^2 \\ \leq& \hat{J}(\boldsymbol{\theta}_{k+1}^f, \boldsymbol{\theta}_{k+1}^J, \boldsymbol{\theta}_k^\mu) -\hat{J}(\boldsymbol{\theta}_{k+1}^f, \boldsymbol{\theta}_{k+1}^J, \boldsymbol{\theta}_{k+1}^\mu) \\&+ \frac{(\alpha_k^\mu)^2 L_{J\theta}^2L_{J\theta\theta}}{2}\\=& \hat{J}(\boldsymbol{\theta}_{k+1}^f, \boldsymbol{\theta}_{k+1}^J, \boldsymbol{\theta}_k^\mu)-\hat{J}(\boldsymbol{\theta}_k^f, \boldsymbol{\theta}_k^J, \boldsymbol{\theta}_k^\mu) + \hat{J}(\boldsymbol{\theta}_k^f, \boldsymbol{\theta}_k^J, \boldsymbol{\theta}_k^\mu) \\ &-\hat{J}(\boldsymbol{\theta}_{k+1}^f, \boldsymbol{\theta}_{k+1}^J, \boldsymbol{\theta}_{k+1}^\mu) + \frac{(\alpha_k^\mu)^2 L_{J\theta}^2L_{J\theta\theta}}{2}.
        \end{aligned}\end{equation}
By applying the Taylor expansion to \eqref{eq_thm2_2} and considering linear approximators $\boldsymbol{g}$ and $V$, the following expression is obtained:
\begin{equation}\label{eq_thm2_eq4}
\begin{aligned} &\alpha_k^\mu\parallel\nabla_{\boldsymbol{\theta}^\mu}\hat{J}(\boldsymbol{\theta}_{k+1}^f, \boldsymbol{\theta}_{k+1}^J, \boldsymbol{\theta}_k^\mu) \parallel^2 - \frac{(\alpha_k^\mu)^2 L_{J\theta}^2L_{J\theta\theta}}{2}  \\ \leq &\hat{J}(\boldsymbol{\theta}_k^f, \boldsymbol{\theta}_k^J, \boldsymbol{\theta}_k^\mu)-\hat{J}(\boldsymbol{\theta}_{k+1}^f, \boldsymbol{\theta}_{k+1}^J, \boldsymbol{\theta}_{k+1}^\mu) \\ &+\langle \nabla_{\boldsymbol{\theta}^f}\hat{J}(\boldsymbol{\theta}_k^f, \boldsymbol{\theta}_k^J, \boldsymbol{\theta}_k^\mu), \boldsymbol{\theta}_{k+1}^f -\boldsymbol{\theta}_k^f\rangle \\&+ \langle \nabla_{\boldsymbol{\theta}^J}\hat{J}(\boldsymbol{\theta}_k^f, \boldsymbol{\theta}_k^J, \boldsymbol{\theta}_k^\mu), \boldsymbol{\theta}_{k+1}^J -\boldsymbol{\theta}_k^J\rangle.
        \end{aligned}\end{equation} Here, to simplify the notation, we denote $\nabla_{\boldsymbol{\theta}^\mu}\hat{J}(\boldsymbol{\theta}_k^\mu) = \nabla_{\boldsymbol{\theta}^\mu}\hat{J}(\boldsymbol{\theta}_{k+1}^f, \boldsymbol{\theta}_{k+1}^J,\boldsymbol{\theta}_k^\mu)$, $\nabla_{\boldsymbol{\theta}^f}\hat{J}(\boldsymbol{\theta}_k^f)=\nabla_{\boldsymbol{\theta}^f}\hat{J}(\boldsymbol{\theta}_k^f, \boldsymbol{\theta}_k^J, \boldsymbol{\theta}_k^\mu)$ and $\nabla_{\boldsymbol{\theta}^J}\hat{J}(\boldsymbol{\theta}_k^J) = \nabla_{\boldsymbol{\theta}^J}\hat{J}(\boldsymbol{\theta}_k^f, \boldsymbol{\theta}_k^J, \boldsymbol{\theta}_k^\mu)$ in the rest of this proof. Then taking the summation over the iteration steps $k=0,1,2,\cdots,K-1$ on both sides of \eqref{eq_thm2_eq4} and applying Cauchy–Schwarz inequality yields: 
\begin{equation}\label{eq_thm2_eq5}
\begin{aligned} &\sum_{k=0}^{K-1}\alpha_k^\mu\parallel\nabla_{\boldsymbol{\theta}^\mu}\hat{J}(\boldsymbol{\theta}_k^\mu) \parallel^2 - \frac{\sum_{k=0}^{K-1}(\alpha_k^\mu)^2 L_{J\theta}^2L_{J\theta\theta}}{2}
\\ \leq& \sum_{k=0}^{K-1}\parallel\boldsymbol{\theta}_{k+1}^f -\boldsymbol{\theta}_k^f \parallel \parallel\nabla_{\boldsymbol{\theta}^f}\hat{J}(\boldsymbol{\theta}_k^f)\parallel  + \parallel\boldsymbol{\theta}_{k+1}^J -\boldsymbol{\theta}_k^J\parallel\\&\times \parallel\nabla_{\boldsymbol{\theta}^J}\hat{J}(\boldsymbol{\theta}_k^J)\parallel + \hat{J}( \boldsymbol{\theta}_0^f, \boldsymbol{\theta}_0^J, \boldsymbol{\theta}_0^\mu)-\hat{J}(\boldsymbol{\theta}_K^f, \boldsymbol{\theta}_K^J, \boldsymbol{\theta}_K^\mu)   \\&\\=&  \hat{J}(\boldsymbol{\theta}_0^f, \boldsymbol{\theta}_0^J, \boldsymbol{\theta}_0^\mu)  -\hat{J}(\boldsymbol{\theta}^{f*}, \boldsymbol{\theta}^{J*}, \boldsymbol{\theta}^{\mu *}) + \hat{J}(\boldsymbol{\theta}^{f*}, \boldsymbol{\theta}^{J*}, \boldsymbol{\theta}^{\mu *}) \\&-\hat{J}(\boldsymbol{\theta}^{f*}, \boldsymbol{\theta}^{J*}, \boldsymbol{\theta}_K^\mu) \!+\! \hat{J}(\boldsymbol{\theta}^{f*}, \boldsymbol{\theta}^{J*}, \boldsymbol{\theta}_K^\mu) \!-\!\hat{J}(\boldsymbol{\theta}_K^f, \boldsymbol{\theta}_K^J, \boldsymbol{\theta}_K^\mu)  \\&+ \sum_{k=0}^{K-1}\parallel\boldsymbol{\theta}_{k+1}^f-\boldsymbol{\theta}^{f*} + \boldsymbol{\theta}^{f*} -\boldsymbol{\theta}_k^f  \parallel \parallel\nabla_{\boldsymbol{\theta}^f}\hat{J}(\boldsymbol{\theta}_k^f)\parallel \\&+ \parallel\boldsymbol{\theta}_{k+1}^J -\boldsymbol{\theta}^{J*} + \boldsymbol{\theta}^{J*} -\boldsymbol{\theta}_k^J\parallel\parallel\nabla_{\boldsymbol{\theta}^J}\hat{J}(\boldsymbol{\theta}_k^J)\parallel.
        \end{aligned}\end{equation}
Let $C_1 = \max\{L_{vx}c_1c_3, 1\}$. Following the triangle inequality, and Lipschitz continuity, we obtain the following bound:
\begin{equation}\label{eq_thm2_eq6}
\begin{aligned} &\sum_{k=0}^{K-1}\alpha_k^\mu\parallel\nabla_{\boldsymbol{\theta}^\mu}\hat{J}(\boldsymbol{\theta}_k^\mu) \parallel^2 - \frac{\sum_{k=0}^{K-1}(\alpha_k^\mu)^2 L_{J\theta}^2L_{J\theta\theta}}{2}
\\ \leq &2 C_1 \sum_{k=0}^{K-1}(\parallel\boldsymbol{\theta}_{k+1}^f-\boldsymbol{\theta}^{f*}\parallel + \parallel\boldsymbol{\theta}_{k+1}^J -\boldsymbol{\theta}^{J*}\parallel) \\&+ C_1 (\parallel \boldsymbol{\theta}_K^J - \boldsymbol{\theta}^{J*}\parallel + \parallel \boldsymbol{\theta}_K^f - \boldsymbol{\theta}^{f*}\parallel) \\&+ \hat{J}(\boldsymbol{\theta}_0^f, \boldsymbol{\theta}_0^J, \boldsymbol{\theta}_0^\mu)  -\hat{J}(\boldsymbol{\theta}^{f*}, \boldsymbol{\theta}^{J*}, \boldsymbol{\theta}^{\mu *}).
        \end{aligned}\end{equation}
By choosing the step sizes as $\alpha_k^f = \frac{\beta_f}{2+k}$ (Lemma~\ref{thm1}) and $\alpha_J = (1-\gamma)/4$ (Lemma~\ref{TD_result}), and recalling the convergence results of $\boldsymbol{\theta}^f$ from \eqref{eq_thm1_results}, we obtain the following results:
\begin{equation}\label{eq_thm2_eq7_0}
\begin{aligned} &\sum_{k=0}^{K-1}\alpha_k^\mu\parallel\nabla_{\boldsymbol{\theta}^\mu}\hat{J}(\boldsymbol{\theta}_k^\mu) \parallel^2 - \frac{\sum_{k=0}^{K-1}(\alpha_k^\mu)^2 L_{J\theta}^2L_{J\theta\theta}}{2} \\ \leq &\hat{J}(\boldsymbol{\theta}_0^f, \boldsymbol{\theta}_0^J, \boldsymbol{\theta}_0^\mu)  -\hat{J}(\boldsymbol{\theta}^{f*}, \boldsymbol{\theta}^{J*}, \boldsymbol{\theta}^{\mu *}) + C_1\Big((1-\nu_J)^{K/2}\\ &\times \parallel\boldsymbol{\theta}_0^J-\boldsymbol{\theta}^{J*}\parallel + \sqrt{\frac{\nu_f}{2+K}} + 2\sum_{k=0}^{K-1}(\sqrt{\frac{\nu_f}{2+k + 1}} \\& + (1-\nu_J)^{(k+1)/2}\parallel\boldsymbol{\theta}_0^J-\boldsymbol{\theta}^{J*}\parallel)\Big) . \end{aligned}\end{equation}
Furthermore, applying the geometric series formula ($\sum_{k=0}^\infty ar^k = \frac{a}{1-r}, |r|<1$) and the inequality $\sum_{k=1}^K k^{-1/2} \approx\int_{1}^K k^{-1/2}\leq 2\sqrt{K}$ to \eqref{eq_thm2_eq7_0}, and then taking the minimum of the left-hand side of \eqref{eq_thm2_eq7_0}, we obtain:
\begin{equation}\label{eq_thm2_pf_final_case2}
\begin{aligned}
        &\min_{k\in\{0,1,2,\cdots,K-1\}}\parallel \nabla_{\boldsymbol{\theta}^\mu}\hat{J}(\boldsymbol{\theta}_k^\mu) \parallel^2  \\ \leq &\frac{C_1}{\sum_{k=0}^{K-1}\alpha_k^\mu}\Big ( (\frac{2}{1-\sqrt{1-\nu_J}} + (1-\nu_J)^{K/2})\parallel\boldsymbol{\theta}_0^J -\boldsymbol{\theta}^{J*}\parallel \\&+ \sqrt{\nu_f}(K^{-1/2} + 4K^{1/2})+\frac{1}{C_1}(\hat{J}(\boldsymbol{\theta}_0^f, \boldsymbol{\theta}_0^J, \boldsymbol{\theta}_0^\mu)  \\&-\hat{J}(\boldsymbol{\theta}^{f*}, \boldsymbol{\theta}^{J*}, \boldsymbol{\theta}^{\mu *}))\Big) + \frac{\sum_{k=0}^{K-1}(\alpha_k^\mu)^2 L_{J\theta}^2L_{J\theta\theta}}{2\sum_{k=0}^{K-1}\alpha_k^\mu}. 
\end{aligned}
\end{equation}
To achieve $\lim_{k\rightarrow\infty}\parallel \nabla_{\boldsymbol{\theta}^\mu}\hat{J}(\boldsymbol{\theta}_k^\mu) \parallel^2 = 0$, one needs to choose $\sum_{k=0}^\infty \alpha_k^\mu=\infty$ and $\sum_{k=0}^\infty(\alpha_k^\mu)^2<\infty$.

Choosing the step size $\alpha_k^\mu = (k+1)^{-1/4}$ for \eqref{eq_thm2_pf_final_case2}, where $\sum_{k=0}^{K-1}\alpha_k^\mu\approx\int_{k=1}^K k^{-1/4} \leq \frac{4 K^{3/4}}{3}$ and $\sum_{k=0}^{K-1}(\alpha_k^\mu)^2 \approx \int_{1}^K k^{-1/2}\leq 2\sqrt{K}$, then \eqref{eq_thm2_pf_final_case2} becomes 
\begin{equation}\label{eq_thm2_pf_final}
\begin{aligned}
        &\min_{k\in\{0,1,2,\cdots,K-1\}}\parallel \nabla_{\boldsymbol{\theta}^\mu}\hat{J}(\boldsymbol{\theta}_k^\mu) \parallel^2  \\ \leq & 3K^{-1/4}(C_1 \sqrt{\nu_f}+\frac{ L_{J\theta}^2L_{J\theta\theta} }{4}) +\frac{3C_1 \sqrt{\nu_f}K^{-5/4}}{4}  \\& +\! \frac{3K^{-3/4}}{4}\Big(C_1(\frac{2}{1-\sqrt{1-\nu_J}} + (1-\nu_J)^{K/2})\parallel\!\boldsymbol{\theta}_0^J -\boldsymbol{\theta}^{J*}\!\!\parallel \\&+\hat{J}(\boldsymbol{\theta}_0^f, \boldsymbol{\theta}_0^J, \boldsymbol{\theta}_0^\mu) -\hat{J}(\boldsymbol{\theta}^{f*}, \boldsymbol{\theta}^{J*}, \boldsymbol{\theta}^{\mu *}) \Big)\\ \eqqcolon&L_a K^{-1/4} + L_b K^{-3/4} +L_c K^{-5/4}, \nonumber
\end{aligned}
\end{equation} 
where \begin{equation}
    \begin{aligned}
        L_a &= 3(C_1 \sqrt{\nu_f}+\frac{ L_{J\theta}^2L_{J\theta\theta} }{4}), \\L_b &= \frac{3}{4}\Big(C_1(\frac{2}{1-\sqrt{1-\nu_J}} + (1-\nu_J)^{K/2})\parallel\boldsymbol{\theta}_0^J -\boldsymbol{\theta}^{J*}\parallel \\& \quad +\hat{J}(\boldsymbol{\theta}_0^f, \boldsymbol{\theta}_0^J, \boldsymbol{\theta}_0^\mu)  -\hat{J}(\boldsymbol{\theta}^{f*}, \boldsymbol{\theta}^{J*}, \boldsymbol{\theta}^{\mu *})\Big),\\ L_c & = \frac{3C_1 \sqrt{\nu_f}}{4}.
    \end{aligned}
\end{equation} $\hfill \blacksquare$

\section{References}
\bibliographystyle{unsrt}
\bibliography{refs_hao.bib}

\end{document}